\documentclass[runningheads]{llncs}

\usepackage{eccv}

\usepackage{eccvabbrv}

\usepackage{graphicx}
\usepackage{booktabs}

\usepackage{algorithm}
\usepackage{algpseudocode}

\usepackage[accsupp]{axessibility}  %

\usepackage{hyperref}

\usepackage{orcidlink}
\usepackage[dvipsnames]{xcolor}

\usepackage{graphicx}

\usepackage{amsmath}
\usepackage{multirow}
\usepackage{epstopdf}
\usepackage{caption}
\usepackage{subcaption}
\usepackage{bm}
\usepackage{pgfplots}
\pgfplotsset{compat=1.15}
\usepackage{verbatim}

\usepackage{footnote}
\usepackage{stfloats}
\usepackage{placeins}
\usepackage{dsfont}
\usepackage{pbox}
\usepackage{enumerate}
\usepackage{tabulary}
\usepackage{dsfont}
\usepackage{mathtools}
\usepackage{amssymb}

\tikzstyle{intt}=[draw,text centered,minimum size=6em,text width=5.25cm,text height=0.34cm]
\tikzstyle{intl}=[draw,text centered,minimum size=2em,text width=2.75cm,text height=0.34cm]
\tikzstyle{int}=[draw,minimum size=2.5em,text centered,text width=3.5cm]
\tikzstyle{intg}=[draw,minimum size=3em,text centered,text width=6.cm]
\tikzstyle{sum}=[draw,shape=circle,inner sep=2pt,text centered,node distance=3.5cm]
\tikzstyle{summ}=[drawshape=circle,inner sep=4pt,text centered,node distance=3.cm]

\newcommand{\norm}[1]{\left \lVert #1 \right \rVert}

\usepackage{algpseudocode}

\usepackage[resetlabels,labeled]{multibib}
	
\usepackage{color, colortbl}
\usepackage[export]{adjustbox}

\begin{document}

\title{TriNeRFLet: A Wavelet Based Triplane NeRF Representation} 

\titlerunning{TriNeRFLet}

\author{Rajaei Khatib\orcidlink{0000-0002-1376-1840} \and
Raja Giryes\orcidlink{0000-0002-2830-0297} }

\authorrunning{Khatib et al.}

\institute{Tel Aviv University \\
\email{rajaeikhatib@mail.tau.ac.il and raja@tauex.tau.ac.il}}

\maketitle

\begin{abstract}
In recent years, the neural radiance field (NeRF) model has gained popularity due to its ability to recover complex 3D scenes. Following its success, many approaches proposed different NeRF representations in order to further improve both runtime and performance. One such example is Triplane, in which NeRF is represented using three 2D feature planes. This enables easily using existing 2D neural networks in this framework, e.g., to generate the three planes. Despite its advantage, the triplane representation lagged behind in 3D recovery quality compared to NeRF solutions. In this work, we propose the TriNeRFLet framework, where we learn the wavelet representation of the triplane and regularize it. This approach has multiple advantages: (i) it allows information sharing across scales and regularization of high frequencies; (ii) it facilitates performing learning in a multi-scale fashion; and (iii) it provides a `natural' framework for performing NeRF super-resolution (SR), such that the low-resolution wavelet coefficients are computed from the provided low-resolution multi-view images and the high frequencies are acquired under the guidance of a pre-trained 2D diffusion model. We show the SR approach's advantage on both Blender and LLFF datasets.

\end{abstract}
    
\section{Introduction}
\label{sec:intro}

\begin{figure}[t]
    \centering
        \includegraphics[width=0.24\columnwidth]{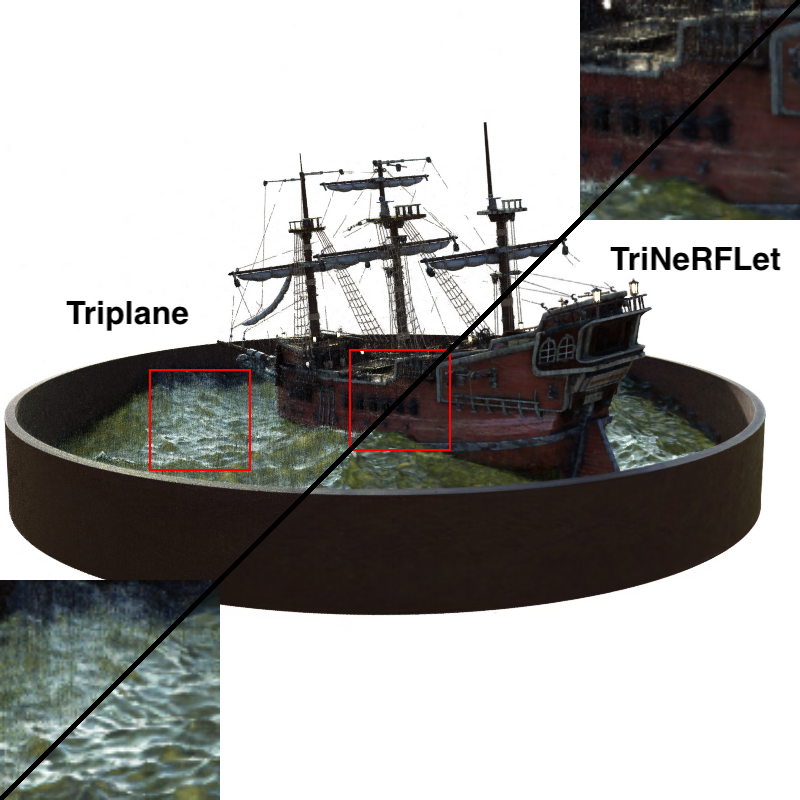}
        \includegraphics[width=0.24\columnwidth]{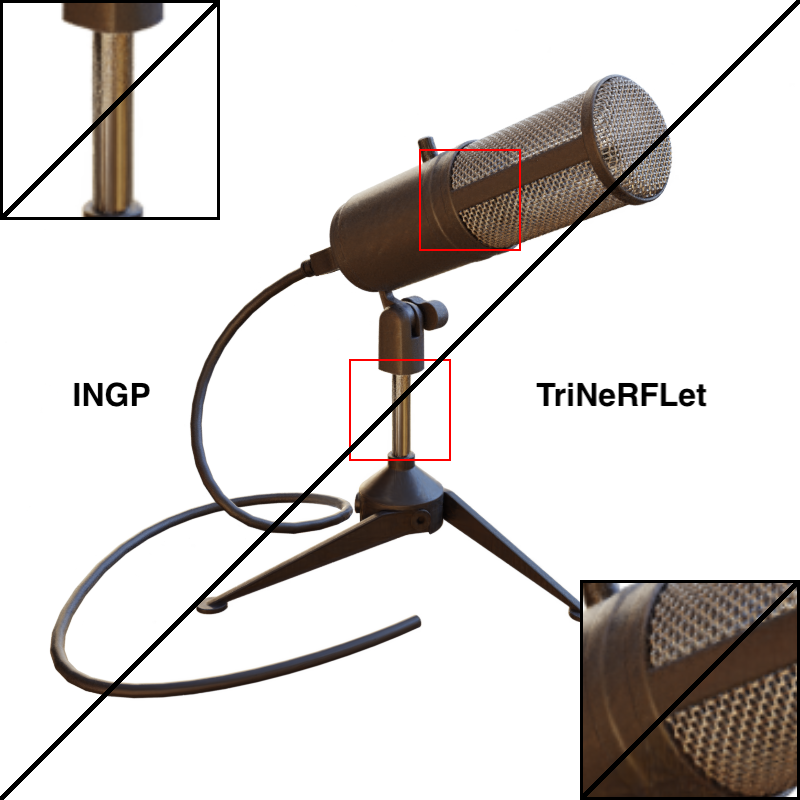}
        \includegraphics[width=0.24\columnwidth]{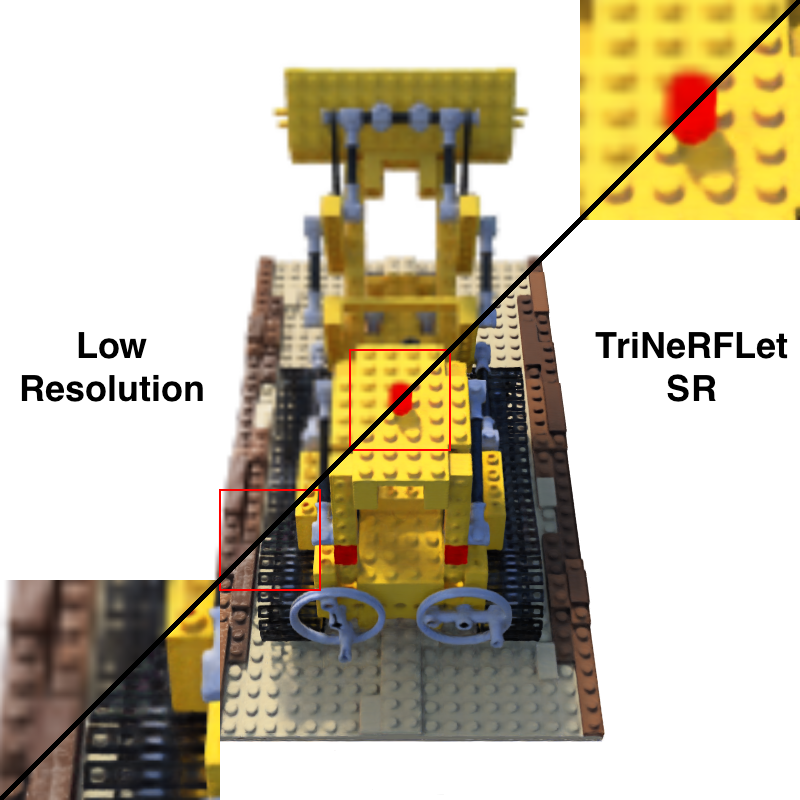}
        \includegraphics[width=0.24\columnwidth]{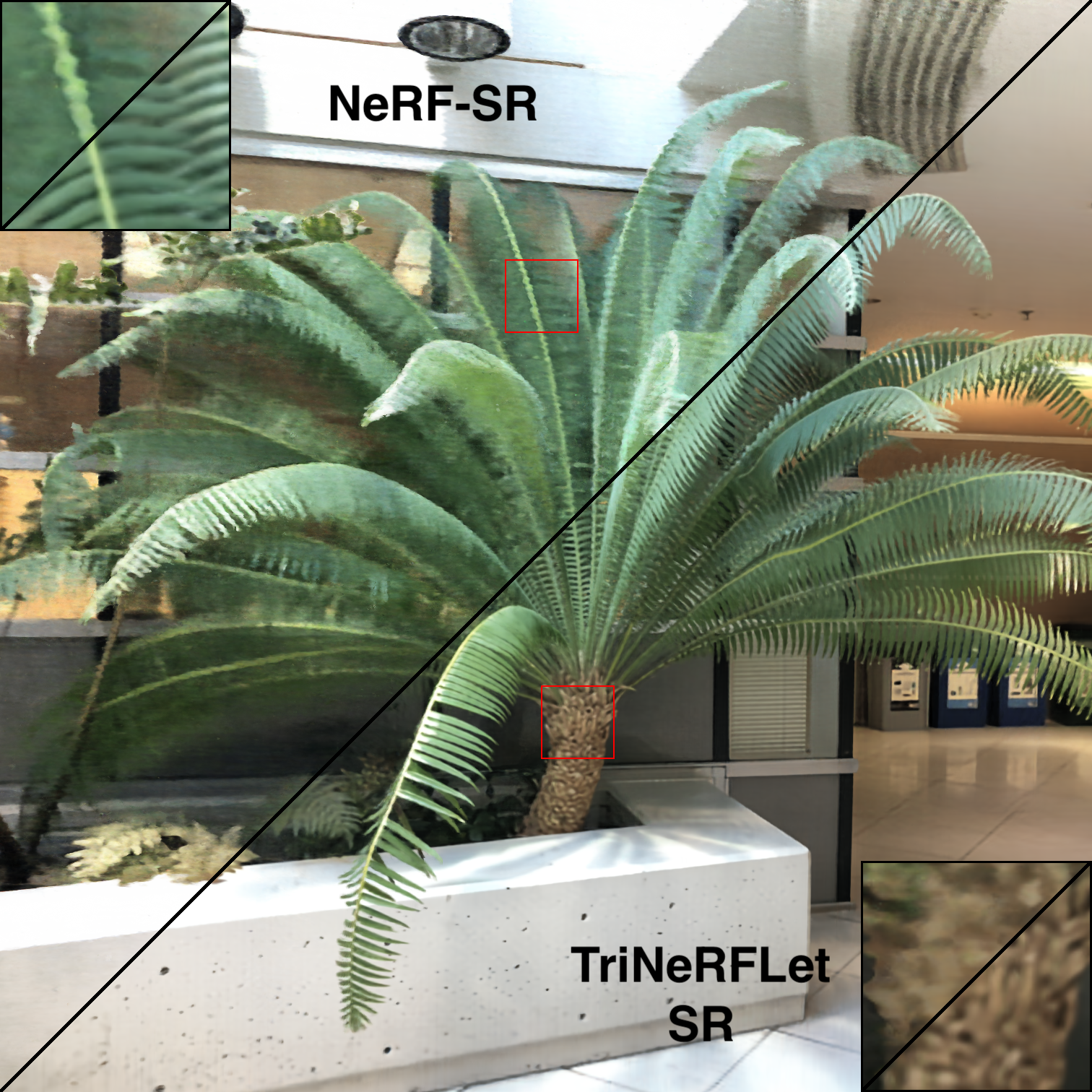}
    \caption{Our approach improves the quality of NeRF reconstruction (zoom-in). From left to right: TriNeRFLet improvement over Triplane, TriNeRFLet compared to INGP, TriNeRFLet SR improvement, TriNeRFLet SR compared to NeRF-SR.
    }
    \label{figure:teaser}
    \vspace{-0.2in}
\end{figure}

3D scene reconstruction from multiple 2D views is a challenging task that has been widely studied, and many methods have been proposed to solve it. Neural radiance field (NeRF) \cite{mildenhall2021nerf} is prominent among these methods as it has demonstrated a high level of generalizability in generating novel views with high quality and consistent lighting. 
NeRF utilizes an implicit representation of the 3D scene in the form of a multi-layer perceptron (MLP). This enables it to capture complex 3D geometry and lighting. 

At a high level, NeRF relies on the rendering equation that approximates each pixel in the image using points sampled along the ray that passes through it \cite{mildenhall2021nerf}. The MLP in NeRF takes as input the frequency encoding of the Euclidean coordinates and view direction of the point of interest, and outputs the radiance and density at this point.
The MLP weights are learned via end-to-end optimization, by comparing the values between the rendered pixel and its value in the corresponding 2D image. After the training process is over, the MLP can be used to render novel views.

Due to its success in representing 3D scenes, many methods proposed improvements to NeRF \cite{barron2021mip,barron2022mip,muller2022instant}. They sought to enhance NeRF 3D reconstruction capabilities, as well as other drawbacks such as high runtime and aliasing artifacts that it suffered from. 
One such approach uses three axis-aligned 2D feature planes, denoted as Triplane, to represent the NeRF \cite{chan2022efficient}. In the rendering process, each point is sampled by projecting it onto each of the three planes and then concatenating the features that correspond to the three projections. This forms a single feature vector for the point that is then passed to a small MLP that outputs the density and color values of this point. 

A significant advantage of the Triplane representation is that it can be used with many already existing 2D methods. In the original work \cite{chan2022efficient}, the authors used an existing 2D Generating Adversarial Network (GAN) architecture to generate its planes. Follow-up works employed the 2D property of the Triplane to perform NeRF super-resolution \cite{bahat2022neural} and 3D generation \cite{shue20233dtriplane,li2023instant3dtriplane}.

While being useful due to its special 2D structure, the reconstruction quality achieved by Triplane lagged behind other efficient multiview reconstruction methods such as instant NGP (INGP) \cite{muller2022instant}, which is an improvement of NeRF, and 3D Gaussian splatting \cite{kerbl20233dsplatting}. Due to its structure, only Triplane entries that are included in the training views rays are learned. Thus, there might be entries at the planes that will still have their initial random values also after training finishes. These random features are used by novel views that rely on them and therefore they deteriorate the quality of their generation and create artifacts in these novel views.

\begin{figure*}[t]
    \centering
    \begin{subfigure}[b]{0.49\textwidth} 
        \centering
        \includegraphics[width=\textwidth]{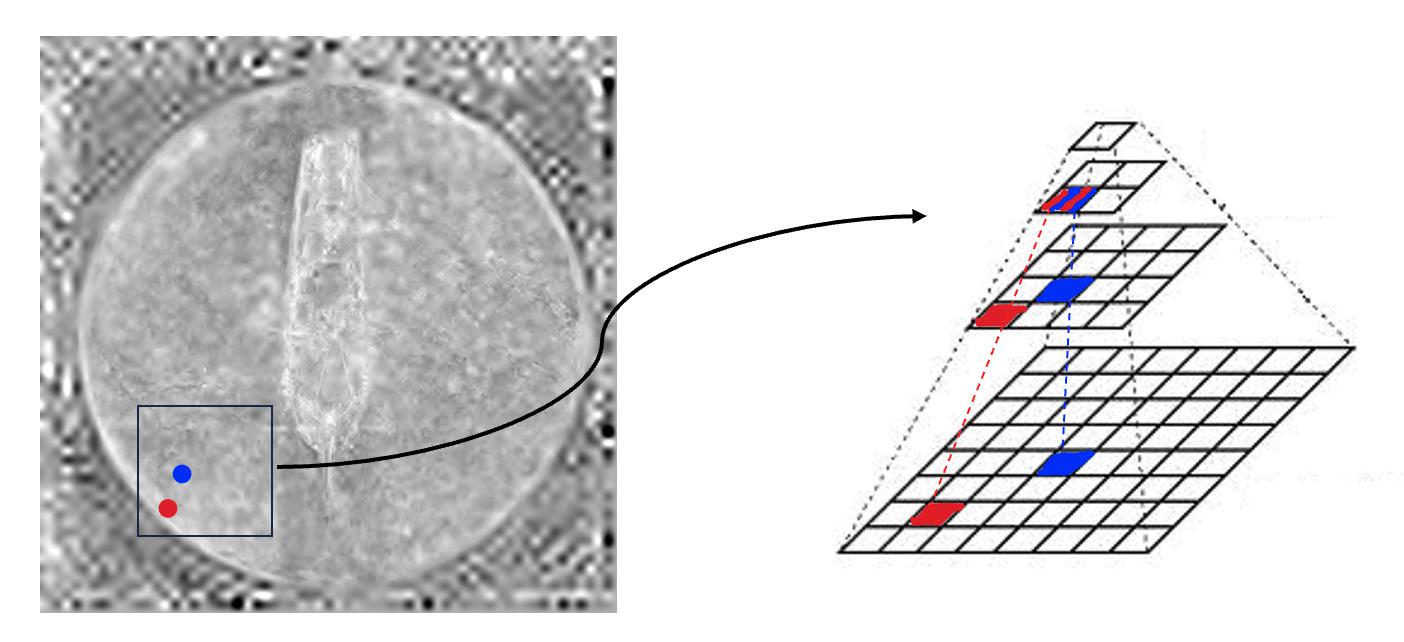}
        \caption{The multiscale wavelet representation transfers information between points on the triplane planes: Although the \textcolor{red}{red} dot on the plane was not covered by the training images views, it will get a coarser estimate from a lower resolution wavelet layer that corresponds also to the \textcolor{blue}{blue} dot, which was updated in training.}
        \label{figure:pyramid} 
    \end{subfigure}\hfill
    \begin{subfigure}[b]{0.49\textwidth} 
        \centering
        \includegraphics[width=\textwidth]{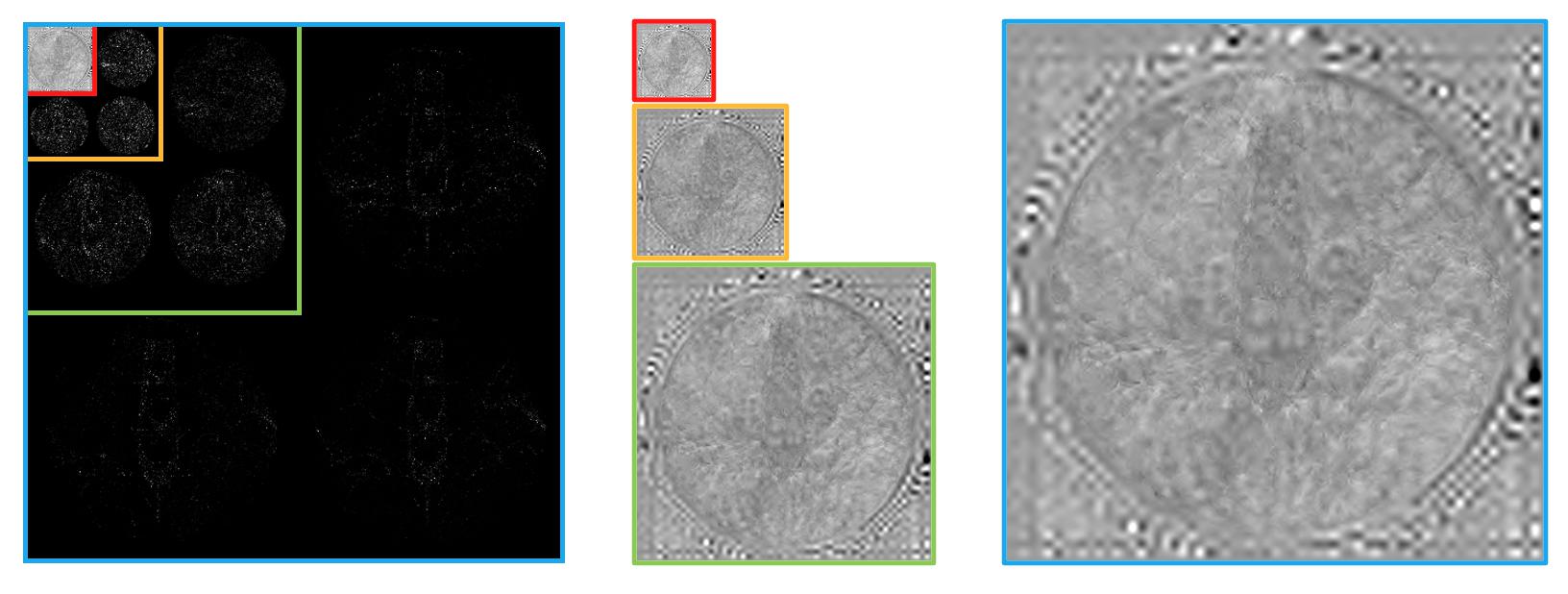}
        \caption{Example of the multiscale property of the wavelet representation. On the left, we present the wavelet representation.
        On the right, we show the feature plane of the wavelet representation on the left at different resolutions. Note that the wavelet representation of a higher resolution includes the wavelet representations of the lower-resolution planes (the colors correspond to the matching between each plane and its wavelet representation).}
        \label{fig:wavelet_multires_prop}
    \end{subfigure}%
    \caption{Illustartion of the multiscale property of the wavelet representation.}
    \vspace{-0.2in}
     
\end{figure*}

To tackle these drawbacks, we present a novel Triplane representation that relies on a multiscale 2D wavelet structure. Instead of learning the feature planes directly, we learn the wavelet features of several resolutions, while regularizing the wavelet representation to be sparse. Thus, regions that are covered by the training views are learned in a similar way to the case of a regular Triplane, and regions that are not covered by the training views are updated by a lower resolution estimate according to how much they are affected by their neighboring regions. Figure~\ref{figure:teaser}(left) shows the improvement in reconstruction quality attained by this representation. Figure~\ref{figure:pyramid} illustrates how regions on the plane may affect each other due to the wavelet representation. 

In our optimization, we use $L_1$ regularization on the wavelet coefficients. The goal is to sparsify the coefficients that are not trained by the given views such that only the coefficients that were trained will impact the reconstruction also from other views. Using $L_1$ regularization is inspired by the wavelet literature, which indicates that wavelet features can represent data accurately while being sparse \cite{guo2022review}. 

To further improve the training, we perform it in a multi-scale fashion, starting with a lower resolution version that updates only the coarse coefficients of the wavelet. Then, after some iterations, we increase the resolution and add additional wavelet layers to the representation. 

We apply the proposed TriNeRFLet framework to two tasks. First, for 3D reconstruction, where it closes the performance gap of Triplane against other multiview 3D reconstruction methods and makes it competitive with current state-of-the-art (SOTA) methods. Second, we combine it with a pre-trained 2D super-resolution (SR) diffusion models \cite{rombach2022high,saharia2022photorealistic} to perform NeRF SR. Figure~\ref{figure:teaser}(right) demonstrates the improvement this approach attains. We show in various experiments the advantage of our TriNeRFLet in novel view reconstruction compared to INGP and Triplane and the superiority of our SR solution compared to competing approaches that do not require dedicated training for multiview or 3D data.

\begin{figure*}[t]
    \centering
    \begin{subfigure}[b]{0.54\textwidth}
        \begin{subfigure}[b]{0.95\textwidth}
            \includegraphics[width=\textwidth,left]{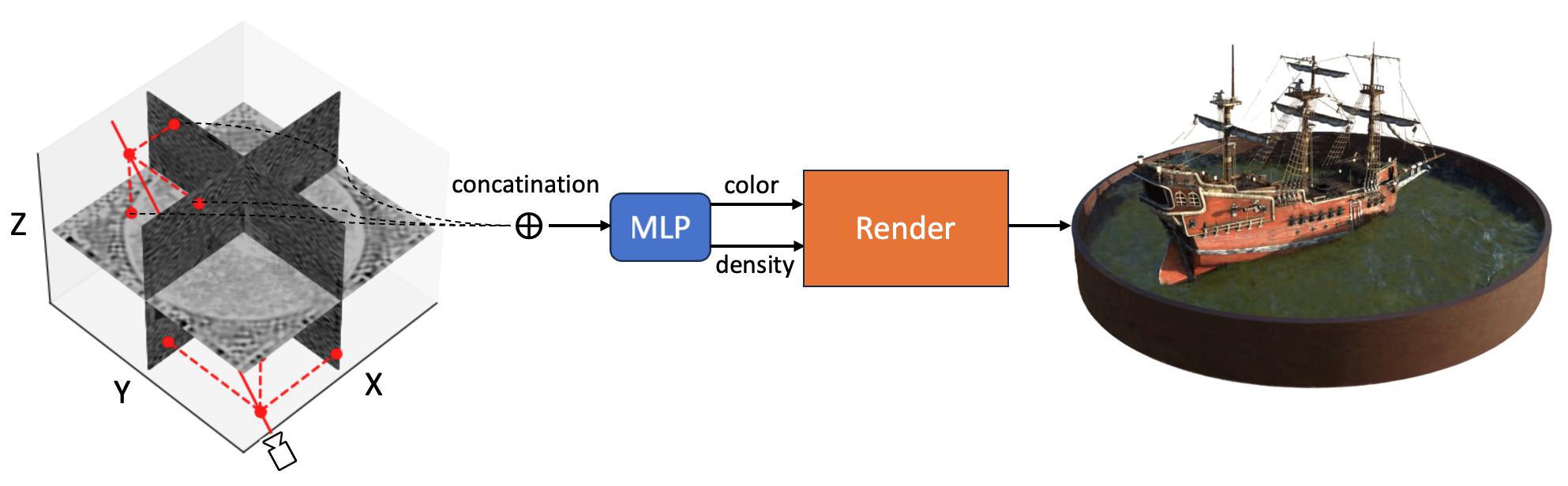}
            \caption{\emph{Triplane rendering.} The feature vector of each point is sampled by gathering the projected features on the Triplane planes. This vector is passed to an MLP to obtain color and density at this point. The final value of a pixel is calculated by using the color and density of the points that reside on the ray that passes through it.}
            \label{subfigure:triplane_render}
        \end{subfigure}
        \begin{subfigure}[b]{0.95\textwidth}
            \includegraphics[width=\textwidth,left]{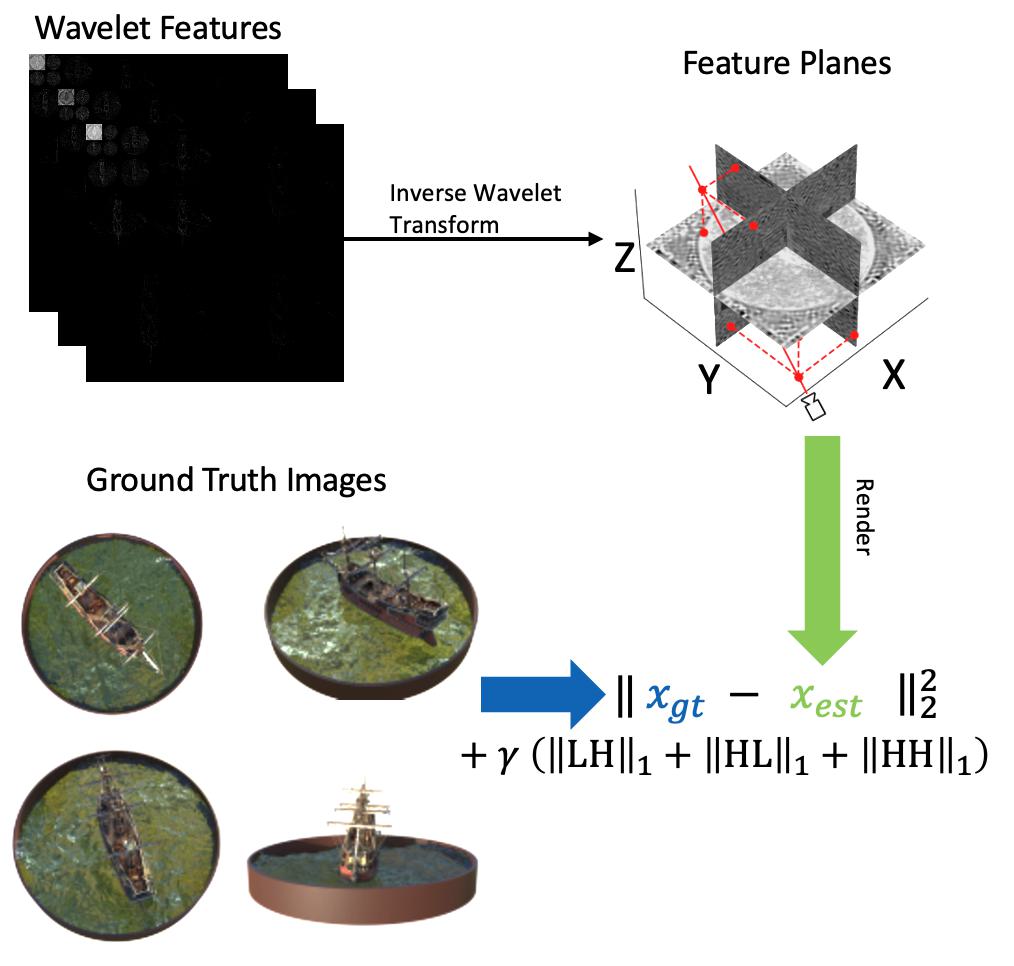}
            \caption{\emph{3D reconstruction training scheme.} First, wavelet features are transformed into the Triplane domain. Next, pixels are rendered using these features in order to fit them to their ground-truth values. The high frequencies channels of LH, HL and HH from all wavelet levels get regularized by $L_1$ loss.}
            \label{subfigure:trinerflet_method}
        \end{subfigure}
    \end{subfigure}
    \begin{subfigure}[b]{0.44\textwidth} 
        \centering
        \includegraphics[width=\textwidth]{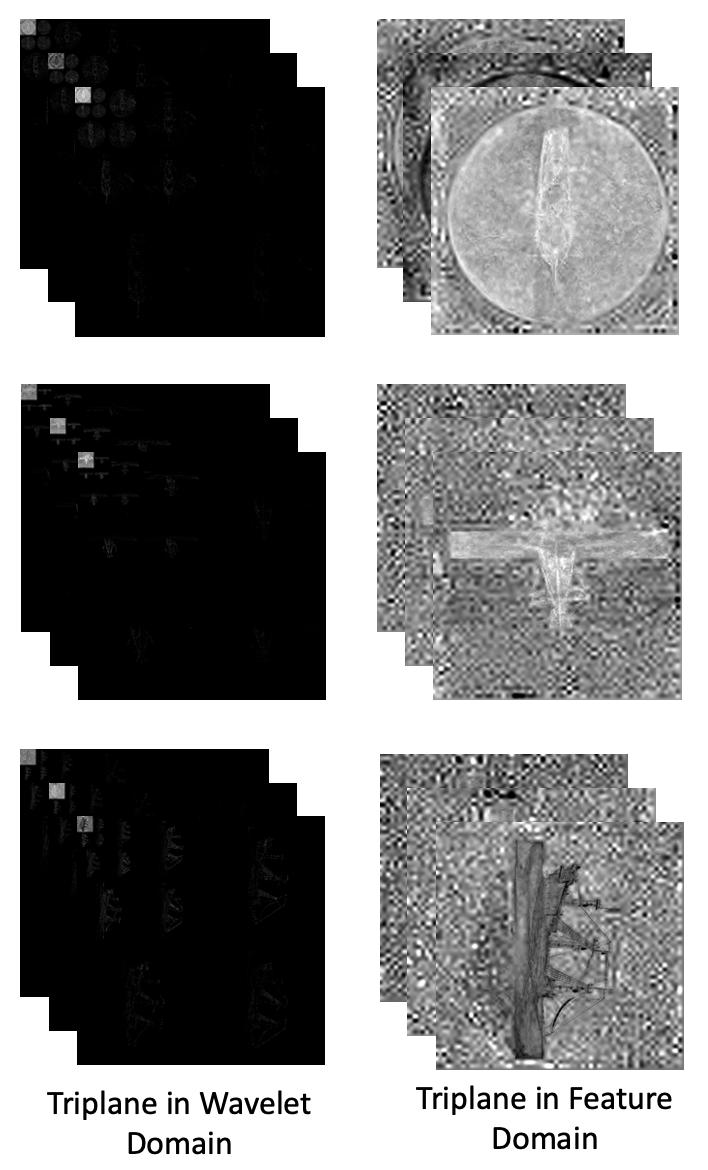}
        \caption{In the wavelet domain, the low-frequency coefficients $\textrm{LL}_{N_{base}}$ are observed at the top left corner (the bright spot). The rest are the high frequencies channels $\textrm{LH}_{N_{base}}$, $\textrm{HL}_{N_{base}}$ and $\textrm{HH}_{N_{base}}$. All the other LH, HL and HH channels can be observed accordingly. All these planes belong to the same 3D object.}
        \label{subfigure:wavelet}
    \end{subfigure}%
     \vspace{-0.05in}
    \caption{\emph{The TriNeRFLet reconstruction framework} learns features in wavelet domain, which are transformed into feature domain to render 3D objects.}
     \vspace{-0.2in}
    \label{figure:scheme}
\end{figure*}

\section{Related Works and Background}
\label{sec:related_works}

\textbf{NeRF} \cite{mildenhall2021nerf} is a framework for reconstructing a 3D scene from a set of multiview images. Its main component is an implicit neural representation of the 3D scene via a mapping neural network ${F_{\theta}} : (\bm{x},\bm{d}) \rightarrow (\bm{c},\sigma)$, where $\bm{x}$ is the Cartesian coordinate of the point of interest, $\bm{d}$ is the direction vector from it to the camera origin, $\bm{c}$ is the RGB color at this point and $\sigma$ is the density. 

The network $F$ is trained using the pixels of the multiview images using the rendering equation. For a given pixel, it relates its value to $N$ points sampled along the ray that passes through this pixel, where the ray direction is derived from the image view. Denoting by $t_1 < t_2 < ... < t_N$ the depth of the points, $\delta_i = t_{i+1} - t_i$ the distance between adjacent samples, and $\sigma_i,\bm{c}_i$ the density and color of each point respectively,
then the pixel value $\hat{C}$ can be estimated from these points via:
\begin{eqnarray}\label{eq:RENDER}
\begin{aligned}
\hat{\bm{C}} = \sum_{i=1}^N T_i (1-\exp(-\sigma_i \delta_i)) \bm{c}_i, 
\quad \textrm{where}  \quad T_i = \exp \left (-\sum_{j=1}^{i-1} \sigma_j \delta_j \right ).
\end{aligned}
\end{eqnarray}
This equation is a discrete approximation of the rendering equation. As we have the true pixel value $\bm{C}$ of the image, the network parameters $\theta$ are trained simply by minimizing the L2 loss $\norm{\bm{C}-\hat{\bm{C}}}_2$.

In the classical NeRF scheme \cite{mildenhall2021nerf}, the mapping function ${F_{\theta}}$ is divided into two main parts. The first part is a frequency encoding function that encodes $\bm{x}$ and $\bm{d}$ using Fourier features \cite{tancik2020fourfeat}. These vectors are then concatenated to form a feature vector that is fed to the second part, which is an MLP function that maps the feature vector into its corresponding color and density values.
Some extensions to NeRF consider optimization using unconstrained photo collections \cite{martinbrualla2020nerfw}, few view points \cite{yu2020pixelnerf}, non-rigidly deforming scenes \cite{park2020nerfies}, and imperfect camera poses~\cite{lin2021barf}.

INGP \cite{muller2022instant} is an important development, which managed to decrease NeRF runtime significantly while getting even better reconstruction performance. It achieved that by using a multiscale hash grid structure alongside a small shallow MLP that represents the scene instead of the big deep one used in previous works. Furthermore, this method introduced new rendering techniques and low-level optimizations in order to get a faster runtime. The multiscale concept of INGP will also be used in our proposed TriNeRFLet framework.

3D Gaussian Splatting \cite{kerbl20233dsplatting} is a state-of-the-art 3D reconstruction approach that takes a different strategy than NeRF. Instead of using an implicit neural function, it represents the scene as a group of 3D Gaussians, and learns their parameters. It introduces a new method for efficiently rendering these Gaussians, thus, managing to train rapidly and achieving 3D rendering in real-time. Note that one needs to distinguish between training and rendering time. While its training time is similar to INGP, its rendering is faster than NeRF solutions.

\noindent \textbf{Triplane} \cite{chan2022efficient,bahat2022neural} is a NeRF variant, in which the frequency encoding part is replaced by three perpendicular 2D feature planes of dimension $N\times N$ with $C$ channels. To get the feature vector of a point $\bm{x}$, the point is projected onto each of these planes, then the sampled feature values of the projected points are gathered together to form a feature vector. The final feature vector is obtained by concatenating this feature vector with the frequency encoding of the direction vector $\bm{d}$. These feature planes are learned alongside the MLP weights with a similar approach to what was explained earlier. Fig.~\ref{subfigure:triplane_render} illustrates this process.

The advantage of Triplane compared to other alternatives is its use of 2D planes, which eases the use of standard 2D networks with it. This has been used in various applications. Bahat et al. \cite{bahat2022neural} utilized it to train a NeRF super-resolution (SR) network from multiview training data. By exploiting the triplane structure, a 2D SR neural network was trained to perform SR on low resolution planes and in this way increase the overall NeRF resolution. Another example is \cite{shue20233dtriplane}, in which a diffusion model was trained to generate a Triplane. It utilizes the generalization power of diffusion models and brings them to 3D generation. Similarly, Triplane is predicted by a transformer-based neural network in \cite{li2023instant3dtriplane} in order to have fast text to 3D generation.

\noindent \textbf{Wavelets} were widely used in signal and image processing applications \cite{Mallat2008Wavelet}. With the advent of deep learning, they were employed in various architectures to improve performance and efficiency. 
The filter-bank approach of wavelet was utilized to represent each kernel in a neural network as a combination of filters from a given basis, e.g., a wavelet basis \cite{qiu2018dcfnet}. 
The concept of using multi-resolution was used to improve positional encoding of implicit representations \cite{hertz2021sape,lindell2021bacon,saragadam2023wire}.
Wavelets were used to improve efficiency and generation of StyleGAN \cite{gal2021swagan}, diffusion models \cite{Guth2022Wavelet,kadkhodaie2023learning,phung2023wavelet} and normalizing flows \cite{jjyu2020waveletflow}.

Rho \emph{et al.} \cite{rho2023masked} implemented a multiscale 3D grid using a grid decomposition method that transforms 2D planes alongside 1D vectors (unrelated to the 2D planes) into a 3D grid, where they used a 2D multiscale wavelet representation to represent the 2D planes. Despite the use of 2D multiscale representation, this method is different in essence from TriNeRFLet, which deploys the 2D wavelet multiscale representation on Triplane planes directly, thus being freed from 3D grid limitations.

\noindent \textbf{Diffusion Models} \cite{dhariwal2021diffusion1, ho2020denoising2} have become a very popular tool in recent years for image manipulation and reconstruction tasks \cite{rombach2022high, kawar2022imagic, kawar2022denoising, whang2022deblurring, SR3}, where a denoising network is trained to learn the prior distribution of the data. Diffusing models can be used also beyond generation, where given a degraded image, some conditioning mechanism is combined with the learned prior to solve different tasks \cite{avrahami2022blended, avrahami2022blendedLatent, Chung2023MR}. For example, in \cite{whang2022deblurring,SR3,abu2022adir} diffusion models were utilized for the problems of deblurring and SR. In this work, we use the Stable Diffusion (SD) upscaler \cite{rombach2022high} to perform SR on 2D projections of the low-resolution NeRF in order to create a high-resolution version of the NeRF. While the SD upscaler is conditioned both on the low-resolution image and a given text prompt, we use it without text, as explained later in the paper.

The authors of \cite{wu2024reconfusion} proposed to fine-tune a diffusion model on the views of a given scene to regularize unseen views. In a similar spirit, we use images generated by a diffusion model to fit the scene and thus improve NeRF SR.

\section{TriNeRFLet Framework}
\label{sec:trinerflet_method}

To alleviate the Triplane drawbacks mentioned above, we present TriNeRFLet. Instead of using the 2D feature planes directly, we optimize them in the wavelet domain. At a high level, wavelet transforms a 2D plane to a multiscale representation that contains both low and high frequencies at different resolutions. The inverse wavelet transform, convert this representation back to the original 2D plane. Note that we apply the wavelet separably on each channel. 
We use the following notation for the wavelet representation. At resolution $i$ of the wavelet representation, we denote by $\textrm{LL}_{i}$ the low frequencies of this resolution, and by $\textrm{LH}_{i}$, $\textrm{HL}_{i}$ and $\textrm{HH}_{i}$ the high frequencies. Note that $\textrm{LL}_i$ is used to create the coefficients of the next resolution of the wavelet representation. At a certain step, we stop and remain with LL.  

In rendering, TriNeRFLet is similar to regular Triplane, in the sense that given a point it calculate its features by projecting it to the Triplane and then feed the features to a MLP that outputs the color $c_i$ and density $\sigma_i$ of the point that are then used by the rendering equation (see Figure \ref{subfigure:triplane_render}). Yet, instead of holding the Triplane in the feature domain, TriNeRFLet holds it in the Wavelet domain and optimizes the wavelet coefficients.
Note that in the regular Triplane scheme, each plane entry impacts only one location in each plane. In TriNeRFLet, it impacts different resolutions in the wavelent representation of the plane (see Figure~\ref{figure:pyramid}).  
Figure \ref{subfigure:wavelet} shows some learned planes for the ship object and their corresponding Wavelet coefficients.

The learned parameters of TriNeRFLet, alongside the MLP weights, are the wavelet coefficients, and they are learned in an end-to-end fashion. The multiscale structure of the wavelet, allows the method to learn fine details whenever it needs, and to have a lower resolution estimate from lower resolution wavelet layers that are affected by nearby values when the area is not covered by the training pixels. The question remains, what should be done with the coefficients of the higher frequencies that are not updated. 

\noindent \textbf{Wavelet regularization.} From wavelet theory \cite{guo2022review}, we expect the LH, HL and HH channels to be sparse. Thus, we regularize these channels to be sparse over all resolutions by adding an $L_1$ regularization for these coefficients to the training loss.
This allows us to increase the resolution of the planes without having the problem of having features in it that remain random or redundant. In TriNeRFLet, also these features get updated.

To further comprehend the significance of the multiscale wavelet structure and the regularization we use for it, we compare the behavior to rendering novel views with a regular Triplane. In a regular Triplane, some features will still have their initial random values after training ends. This introduces artifacts when rendering novel views, as demonstrated in the ship example in Figure \ref{figure:teaser}. Note that these artifacts are absent in TriNeRFLet because all the features in the planes are updated either by direct training or using a coarser estimate from lower resolution wavelet coefficients (see Figure \ref{figure:pyramid}).

\noindent \textbf{Multiscale training.} A key advantage of TriNeRFLet is that the multiscale property of the wavelet representation enables increasing the resolution of the TriNeRFLet during training seamlessly. It is possible to continue to train the higher-resolution TriNeRFLet while using the wavelet coefficients of the lower-resolution TriNeRFLet. The wavelet of the lower-resolution plane is a subset of the wavelet of the higher resolution plane (see Figure~\ref{fig:wavelet_multires_prop}). We use this flexibility to improve the training by learning the wavelet planes in a coarse to fine manner. As we explain later, we leverage the multiscale structure of the wavelet both for NeRF reconstruction and SR.

The multiscale wavelet structure is not ideal in terms of training runtime due to the fact that the feature planes need to be reconstructed in each training step. Yet, its rendering runtime is quite fast since the feature planes are reconstructed just once and then the runtime is competitive with other well-known fast NeRF schemes like INGP \cite{muller2022instant}.

\section{TriNeRFLet Applications}
\subsection{TriNeRFLet Reconstruction}
\label{sec:trinerflet_recon}

The TriNeRFLet framework is straightforwardly applied to 3D scene reconstruction task. Given the training pixels, wavelet features are optimized end-to-end on the pixel reconstruction loss in addition to the wavelet high frequencies regularization (see Figure~\ref{subfigure:trinerflet_method}). 
The overall loss is:
\begin{eqnarray*}
\mathcal{L} = \sum_{i \in TrainingPixels} \norm{\bm{C}_i-\hat{\bm{C}_i}}_2^2 \, +  
\gamma \sum_{l \in resolutions} \left( \norm{\textrm{LH}_l}_1 + \norm{\textrm{HL}_l}_1 + \norm{\textrm{HH}_l}_1 \right),
\end{eqnarray*}
where $\gamma$ is the $L_1$ regularization factor.

As mentioned before, the flexible multiscale structure of TriNeRFLet enables the training to be in a coarse-to-fine manner. This mitigates the runtime overhead of the wavelet inverse transform. At the beginning of the training, the method usually learns high-level features, and then when it starts to converge it learns the low-level details, which is coherent with the coarse-to-fine approach.

\subsection{TriNeRFLet Super-Resolution (SR)}
\label{sec:trinerflet_sr}

We turn to present a novel NeRF SR approach that relies on the 
proposed multiscale structure. This method employs a 2D pre-trained stable diffusion (SD) upscaler \cite{rombach2022high} for SR ``guidance''. Our approach does not require any kind of low-high resolution 3D scene pairs for supervision as done in other works \cite{bahat2022neural}. Utilizing the robustness of the SD upscaler, our approach has the potential to handle different types of scenes and a range of resolutions.

Given a set of low-resolution (LR) multiview images of a given scene, the goal is to generate high-resolution (HR) views of this scene. 
We first apply TriNeRFLet using the LR images, providing a LR 3D reconstruction. In this reconstruction, we learn only the low frequencies of the wavelet representation. 
Denote the size of the LR planes by $N_{LR}$ and the number of wavelet levels by $L_{LR}$. For the HR planes, denote the size by $N$ and the number of levels by $L$. Due to the multiscale nature wavelet representations, in the HR TriNeRFLet we just need to learn the high frequencies of the wavelet, i.e., the first $L_{LR}$ levels of the HR wavelet planes are shared with the LR TriNeRFLet reconstruction. 

\begin{algorithm}[t]
\caption{TriNeRFLet Super-Resolution (SR)}\label{alg:trinerflet_sr}
\renewcommand{\algorithmicrequire}{\textbf{Input:}}
\renewcommand{\algorithmicensure}{\textbf{Output:}}
\begin{algorithmic}[1]
\Require Ground truth low resolution (LR) images $\left \{ {\bm{x}_i}^{gt}_{LR} \right \}_{i=1}^k$, LR only steps $s_{LR}$, Total steps $s$, Wavelet LR levels $L_{LR}$, Refresh steps $s_{refresh}$, Diffusion step limits $T_{min} \; , \; T_{max}$.

\Ensure High-Resolution (HR) TriNeRFLet

\State $\text{HR-set} \gets \{\}$ \Comment{empty dictionary}
\For{$itr=0 \, \cdots \, s-1$}
        \State Choose randomly a frame i - ${\bm{x}_i}^{gt}_{LR}$.
        \State \parbox[t]{\dimexpr\linewidth-\algorithmicindent}{%
          Render $ {\bm{x}_i}^{est}_{LR}$ using only first $L_{LR}$ wavelet levels.
        }
        \State $loss \gets \norm {{\bm{x}_i}^{est}_{LR} - {\bm{x}_i}^{gt}_{LR}}_2^2$
        \If{$itr \geq s_{LR}$}
            \State \parbox[t]{\dimexpr\linewidth-\algorithmicindent}{%
              Render $ {\bm{x}_i}^{est}_{HR}$ using all wavelet levels.
            }
            \State \If{$itr \% s_{refresh} == 0$}
                \State $\text{HR-set} \gets \{\}$ \Comment{empty dictionary}
            \EndIf
            
            \State \If{$i \in \text{HR-set}$}
                    \State ${\bm{x}_i}^{gt}_{HR} \gets \text{HR-set}[i]$
                \Else
                    \State $t \sim Rand(T_{min} \; , \; T_{max})$
                    \State \parbox[t]{\dimexpr\linewidth-\algorithmicindent}{%
                      ${\bm{x}_i}^{gt}_{HR} \gets \text{SD}_\text{refine} ( {\bm{x}_i}^{est}_{HR} , {\bm{x}_i}^{gt}_{LR},t  )$
                    }
                    \State $\text{HR-set}[i] \gets {\bm{x}_i}^{gt}_{HR}$
                \EndIf
            \State \parbox[t]{\dimexpr\linewidth-\algorithmicindent}{%
                      $loss \gets loss + \norm {{\bm{x}_i}^{est}_{HR} - {\bm{x}_i}^{gt}_{HR}}_1 + \lambda \; \text{LPIPS}({\bm{x}_i}^{est}_{HR},{\bm{x}_i}^{gt}_{LR})$
                    }
        \EndIf
        \State \parbox[t]{\dimexpr\linewidth-\algorithmicindent}{%
                      $loss \gets loss + \gamma \; \sum_{l} ( \norm{\textrm{LH}_l}_1 + \norm{\textrm{HL}_l}_1 + \norm{\textrm{HH}_l}_1 )$
                    }
        \State $T_{max} \gets scheduler(T_{max},i)$  \Comment{decrease $T_{max}$}
        
\EndFor
\end{algorithmic}
\end{algorithm}

\begin{figure*}[t]
    \centering
    \begin{subfigure}[b]{0.48\textwidth} 
        \centering
        \includegraphics[width=\textwidth]{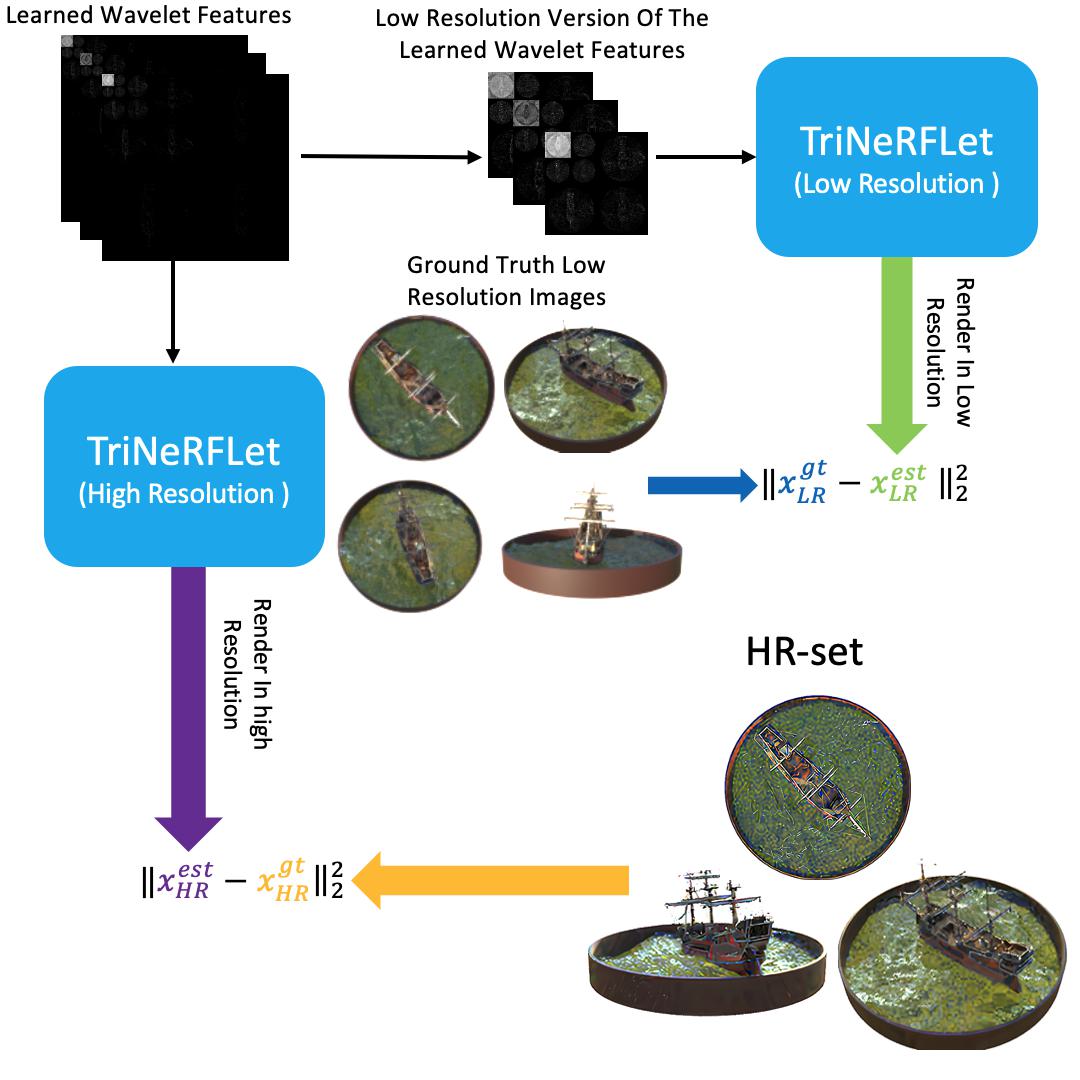}
        \caption{\emph{TriNeRFLet based super-resolution.} Low-resolution TriNeRFLet renders LR images using the LR version of the wavelet features, and fits them to the given low-resolution images. High-resolution TriNeRFLet renders HR images using all wavelet levels, and fits them to HR images from HR-set.}
        \label{figure:sr_method}
    \end{subfigure} \hfill
    \begin{subfigure}[b]{0.48\textwidth} 
        \centering
        \includegraphics[width=\textwidth]{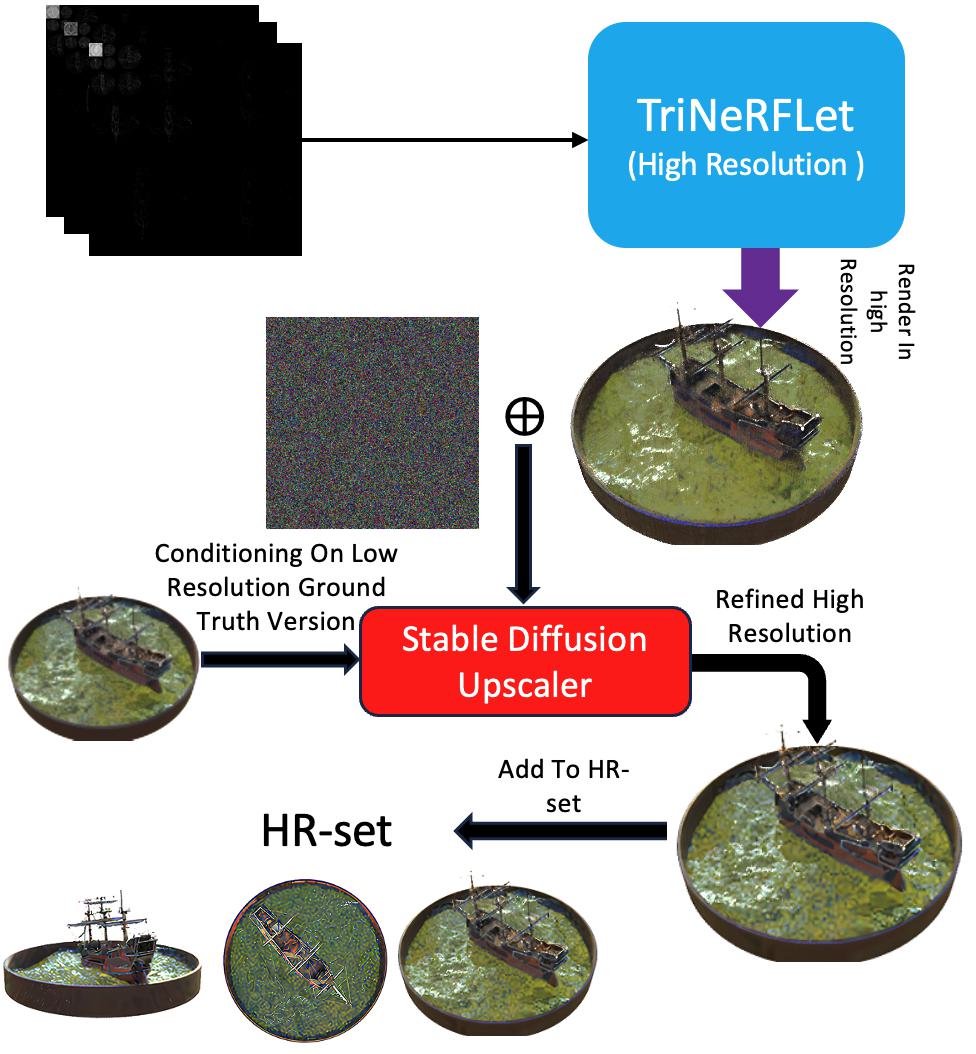}
        \caption{\emph{$\text{SD}_\text{refine}$ process.} First, a HR image is rendered. To improve its quality, noise is added to it and then it is plugged into Stable Diffusion upscaler with its LR version for conditioning. The result, which is a refined HR image is added to the HR-set.}
        \label{figure:hre}
    \end{subfigure}
    \caption{TriNeRFLet Super-Resolution (SR) components.}
    \vspace{-0.2in}
\end{figure*}

Figure \ref{figure:sr_method} illustrates our TriNeRFLetSR strategy. 
First, we start by reconstructng the LR TriNeRFLet using the provided low-resolution multiview images. After training for $s_{LR}$ steps, the super-resolution process starts, and a key component of it is the stable diffusion upscaler refinement ($\text{SD}_\text{refine}$) step presented in Figure~\ref{figure:hre}. Given a low-resolution ground-truth image $\bm{x}^{gt}_{LR}$, its high-resolution version $\bm{x}^{est}_{HR}$ is rendered using all wavelet levels $L$. Initially, the result will not be of high-resolution. Thus, to improve it we use a diffusion step with time $t$ randomly  selected from the range $[T_{min} \; , \; T_{max}]$. Then, similar to \cite{instructnerf2023}, a noise with variance that depends on $t$ is added to $\bm{x}^{est}_{HR}$. Then, we plug the noisy version of $\bm{x}^{est}_{HR}$ into the SD upscaler to perform the diffusion process from step $t$ until the end, while being conditioned on the $LR$ image $\bm{x}^{gt}_{LR}$. This results in the enhanced version of $\bm{x}^{est}_{HR}$, denoted as $\bm{x}^{enhanced}_{HR}$. We denote this enhancement process by $\text{SD}_\text{refine}$.

We use the SD upscaler with a small modification compared to the original work \cite{rombach2022high}. In their work, the denoising step is $$\bm{x}_{t-1} = \epsilon(\bm{x}_t,\bm{z},\varnothing) + \alpha \left ( \epsilon(\bm{x}_t,\bm{z},\bm{y}) - \epsilon(\bm{x}_t,\bm{z},\varnothing) \right),$$ where $\epsilon$, $\alpha$, $\bm{z}$ and $\bm{y}$ are the U-net denoiser, a guidance factor, the low-resolution image and the guidance text respectively. Instead of doing the guidance in the text direction, we do it in the low-resolution image direction and without text, i.e., $\bm{x}_{t-1} = \epsilon(\bm{x}_t,\varnothing,\varnothing) + \alpha \left ( \epsilon(\bm{x}_t,\bm{z},\varnothing) - \epsilon(\bm{x}_t,\varnothing,\varnothing) \right)$.

Having the refined HR images, we want to use them to update the TriNeRFLet. 
We add these images to the set denoted HR-set. This set is initialized to be empty and gets updated sequentially. It is refreshed (i.e. emptied) every $s_{refresh}$ steps as illustrated in Algorithm \ref{alg:trinerflet_sr}. The images in HR-set are used as ``ground-truth'' images for training the HR TriNeRFLet to update all its $L$ levels.

To tie all threads together, the method starts by fitting TriNeRFLet with low-resolution features to the low-resolution ground-truth images. Then $\text{SD}_\text{refine}$ is used to improve the resolution of the TriNeRFLet generated views and the images are stored in HR-set. Next, a TriNeRFLet with high-resolution features (using all wavelet levels) is fitted to HR-set and simultaneously the low-resolution images are used to train the low levels of the wavelet representation of the same TriNeRFLet. As in regular TriNeRFLet training, a $L_1$ regularization of the wavelet coefficients is added to the loss. In addition, we use also a LPIPS loss between the rendered high resolution and the ground truth low resolution. As mentioned earlier, HR-set gets refreshed every $s_{refresh}$ training steps. Furthermore, the diffusion step upper limit $T_{max}$ is scheduled to linearly decrease during training in order to force $\text{SD}_\text{refine}$ to introduce fewer changes as training converges. This is described in Algorithm \ref{alg:trinerflet_sr}.

\begin{table*}[t]
    \resizebox{\textwidth}{!}{\begin{tabular}{ |p{0.2\textwidth} ||p{0.07\textwidth} |p{0.07\textwidth} |p{0.06\textwidth}| p{0.11\textwidth}| p{0.07\textwidth}| p{0.075\textwidth}| p{0.07\textwidth}| p{0.085\textwidth}|p{0.085\textwidth}||p{0.09\textwidth}|p{0.08\textwidth}| }
     \hline
     Method & Mic &Chair& Ship & Materials & Lego & Drums & Ficus & Hotdog & Avg. & Train Time $\downarrow$ & Render FPS $\uparrow$\\
     \hline
     NeRF & $32.91$ &$33.00$& $28.65$ & $29.62$ & $32.54$ & $25.01$ & $30.13$ & $36.18$ & $31.005$ & $\geq 1\text{ day}$ & $\leq 0.2$ \\
     \hline
     INGP (paper)& $36.22$ &$35.00$& $31.10$ & $29.18$ & $36.39$ & $26.02$ & $33.51$ & $37.40$ & $33.176$ & $ 5\text{ mins}$ & $1$ \\
     \hline
     INGP (rerun)& $36.22$ & $34.81$ & $31.1$ & $29.68$ & $36.03$ & $25.15$ & $32.41$ & $37$ & $32.8$ &  $5 \text{ mins}$ & $1$ \\
     \hline
     INGP (rerun)*& $36.3$ & $35.18$ & $31.22$ & $29.8$ & $36.06$ & $24.96$ & $33.22$ & $37.1$ & $32.98$ & $6 \text{hours}$ & $1$ \\
     \hline
     3D Gaussian Splatting& $35.36$ &  $\underline{35.83}$& $30.80$ &  $\underline{30.00}$ & $35.78$ &  $\underline{26.15}$ &  $\textbf{34.87}$ &  $\underline{37.72}$ &$33.32$ & $10\text{ mins}$ & $135$\\
     \hline
      Triplane& $33.85$ &$32.83$& $29.58$ & $28.15$ & $34.70$ & $24.86$ & $30.35$ & $35.80$ & $31.26$ & $2$ hours & $1.5$\\
     \hline
     Triplane*& $34.4$ &$33.5$& $29.97$ & $28.1$ & $35.4$ & $24.9$ & $30.68$ & $36.16$ & $31.64$ & $6$ hours & $1.5$\\
     \hline
     \hline
     Ours: TriNeRFLet Small & $35.18$ &$33.76$& $30.33$ & $29.14$ & $35.32$ & $25.66$ & $33.34$ & $36.24$ & $32.37$ & $15$ mins & $1.5$\\
     \hline
     Ours: TriNeRFLet Base Light& $35.77$ &$35.00$& $31.10$ & $29.35$ & $36.44$ & $25.98$ & $33.96$ & $36.93$ & $33.07$ & $1.5$ hours  & $1.5$\\
     \hline
     Ours: TriNeRFLet Base&  $\underline{36.72}$ &$35.54$& 
$\underline{31.77}$ & $29.72$ &  $\underline{36.66}$ & $26$ & $33.7$ & $37.31$ &  $\underline{33.43}$ & $6$ hours & $1.5$ \\
     \hline
     Ours: TriNeRFLet Large&  $\textbf{37.22}$ &  $\textbf{36}$& $\textbf{32.7}$ &  $\textbf{30.34}$ &  $\textbf{37.32}$ &  $\textbf{26.2}$ & $\underline{34.47}$ & $\textbf{37.89}$ &  $\textbf{34.017}$ & $\sim1$ day & $1.5$ \\
     \hline

    \end{tabular}}
    \caption{Blender Dataset PSNR ($\uparrow$) results with train time and render FPS for each method. \textbf{Bold} is best, \underline{underline} is second. For fair comparison, we also show INGP (rerun)* and Triplane*, which are trained for $6$ hours as our TriNeRFLet base.}
    \label{tab:blender_results}
    \vspace{-0.3in}
\end{table*}

The TriNeRFLet multiscale wavelet structure plays an important role in the SR scheme. First, its multiscale structure establishes a 
connection between high-resolution and low-resolution renderings by forcing them to share information (using the same first $L_{LR} $ wavelet levels). Second, wavelet regularization, especially in the layers that only the HR TriNeRFLet uses, constrains the optimization process and forces the HR TriNeRFLet to learn only useful wavelet coefficients that contribute to HR details. Thus, the combination of multiscale wavelet and diffusion models allows the method to learn the high-resolution details it exactly needs without redundancy.

The diffusion SR model that we use supports SR from $128$ to $512$. 
Yet, we want to support other resolutions, such as $100$ to $400$ or $200$ to $800$. For lower resolutions, e.g., going from $100$ to $400$, we pad $\bm{x}^{gt}_{LR}$ and $\bm{x}^{est}_{HR}$ with zeros in order to achieve the desired resolution before plugging them into the SD upscaler. When it finishes, the result is cropped to the original resolution. For higher resolutions, e.g., going from $200$ to $800$, we randomly crop $128$ and $512$ resolution crops from $\bm{x}^{gt}_{LR}$ and $\bm{x}^{est}_{HR}$, respectively, that correspond to the same relative location. We plug the images into the upscaler, and the result is actually an enhanced version of the crop from $\bm{x}^{est}_{HR}$. Then, we only fit the HR render to this crop instead of the whole image. When the HR-set is refreshed, a different random crop will be chosen, thus covering more areas as the process continues.

\definecolor{lightblue}{rgb}{0.5,0.75,1}
\definecolor{lightgreen}{rgb}{0.5,1,0.5}
\definecolor{LightRed}{rgb}{1,0.41,0.33}

\section{Experiments}
\label{sec:imp_and_results}

We turn to present the results of our method. More visual results and ablations appear in the supplementary material. Code is available in the \href{https://github.com/RajaeeKh/TriNerfLet}{github} repo.

\subsection{3D Scene Reconstruction}
\label{subsec:scene_recovery}
For evaluating TriNeRFLet in the 3D recovery task, we define four versions of TriNeRFLet - small, base light, base and large. The parameters of TriNeRFLet are wavelet LL component resolution - $N_{LL}$, base resolution - $N_{base}$, wavelet layers - $L$, final resolution - $N_{final}$, number of channels - $C$, wavelet regularization $\gamma$, MLP dim (neurons) - $W$, MLP hidden layers - $D_{density}$ and $D_{color}$ and training steps. These parameters are provided for each model in the Supplementary Materials. In the small, base light and base versions, TriNeRFLet uses the exact same MLPs as in INGP. TriNerfLet is trained using the Adam optimizer with a learning rate of $0.01$ and an exponential decay scheduler. The wavelet low frequencies in $\textrm{LL}$ (of size $N_{LL} \times N_{LL}$) are randomly initialized, and the other frequencies in LH, HL and HH at all levels are initialized with zeros. We use the Biorthogonal $6.8$ ($\text{Bior}6.8$) wavelet type. Ablation for other wavelet types is found in the sup. mat. The other training details are similar to INGP. Our implementation uses \cite{torch-ngp}, which is a Pytorch-based implementation of INGP.

Since the wavelet reconstruction component affects the runtime of TriNeRFLet, we start the training with lower resolutions of the planes, training coarse to fine. The initial plane resolution is set to $N_{base}$ and it is increased throughout the training till it reaches the size $N_{final}$ that has $L$ wavelet levels. As described above, due to the wavelet multiscale structure, when we move to the next resolution in training we simply add another level (of higher frequencies) to the wavelet representation and add it to the training, where the lower levels are kept as is (but continue to train). As mentioned in Sec. \ref{sec:trinerflet_method}, the relatively high training runtime is one of TriNeRFLet framework disadvantages, and this is due to the inverse wavelet transform overhead during optimization. Thus, doing the optimization in this coarse to fine manner mitigates this drawback and accelerates the training (inverse wavelet on smaller planes is faster).

To evaluate the performance of our method, we use the Blender dataset \cite{mildenhall2021nerf}, which contains $8$ synthetic 3D scenes with complex geometry and lighting. We compare TriNeRFLet with the vanilla NeRF \cite{mildenhall2021nerf}, regular Triplane (with base config), INGP \cite{muller2022instant} and 3D Gaussian Splatting \cite{kerbl20233dsplatting}. Results are reported in Table \ref{tab:blender_results}.
For INGP we report 3 versions: INGP (paper) denotes the officially published results by \cite{muller2022instant}, and INGP (rerun) and INGP (rerun)* denote the results that we obtained by running INGP (using \cite{muller2022instant} software) for $5$ mins and $6$ hours respectively. Also, Triplane and Triplane* denote Triplane results when trained for $5$ mins and $6$ hours respectively.
We ran all TriNeRFLet versions on a single A6000 GPU. The reported results indeed demonstrate the improvement in performance that the multiscale wavelet brings, as it outperforms regular Triplane by a significant margin, and improves over the current SOTA, even in case they were trained for a longer time.

Regarding rendering time, the TriNeRFLet (all versions) frames per second (FPS) are approximately $1.5$, while INGP and 3D Gaussian Splatting are approximately $1$ and $135$ respectively. These FPS are tested on a single A6000 GPU. Note that TriNeRFLet rendering FPS is compatible and even slightly better than INGP, which is known to be fast within the NeRF approaches. 3D Gaussian Splatting, which is not a NeRF method, is faster in its rendering time since it uses a different internal structure than NeRF. When compared only to NeRF-based solutions, our method is very fast with competitive reconstruction performance. While our rendering time is faster, the training time of TriNeRFLet is higher than INGP. This is due to the wavelet reconstruction, which needs to be done at each training step but only once during rendering time.

\subsection{3D Scene Super-Resolution}
\label{subsec:scene_super_res}

\noindent \textbf{Blender.} We turn to present our results for TriNeRFLet SR.
The objective here is to generate novel views with $\times 4$ resolution than the input images. To this end, we tested two different settings on the Blender dataset \cite{mildenhall2021nerf}, $100$ to $400$ and $200$ to $800$. For the first one, we use $N_{LR} = 256$ and $N = 1024$, while in the second setting $N_{LR} = 512$ and $N = 2048$. Further technical details appear in the Supplementary Materials.

We compare our method with NeRF*, NeRF-Bi, NeRF-Liif, NeRF-Swin, NeRF-SR (SS) (these methods do not require 3D supervision and are described in \cite{wang2022nerf_sr}) and NVSR (results of $200\rightarrow 800$ were provided to us by the authors) \cite{bahat2022neural}. NeRF* is regular NeRF trained on LR views and then rendered in HR resolution, NeRF-Bi, NeRF-Liif and NeRF-Swin are trained on LR views, then rendered in LR resolution and upscaled using different 2D SR methods \cite{wang2022nerf_sr}. 

\begin{figure}[t]
\centering
    \begin{tabular}{   p{0.24\textwidth}  p{0.24\textwidth} p{0.24\textwidth}  p{0.24\textwidth}}

    \includegraphics[width=0.248\textwidth,valign=m]{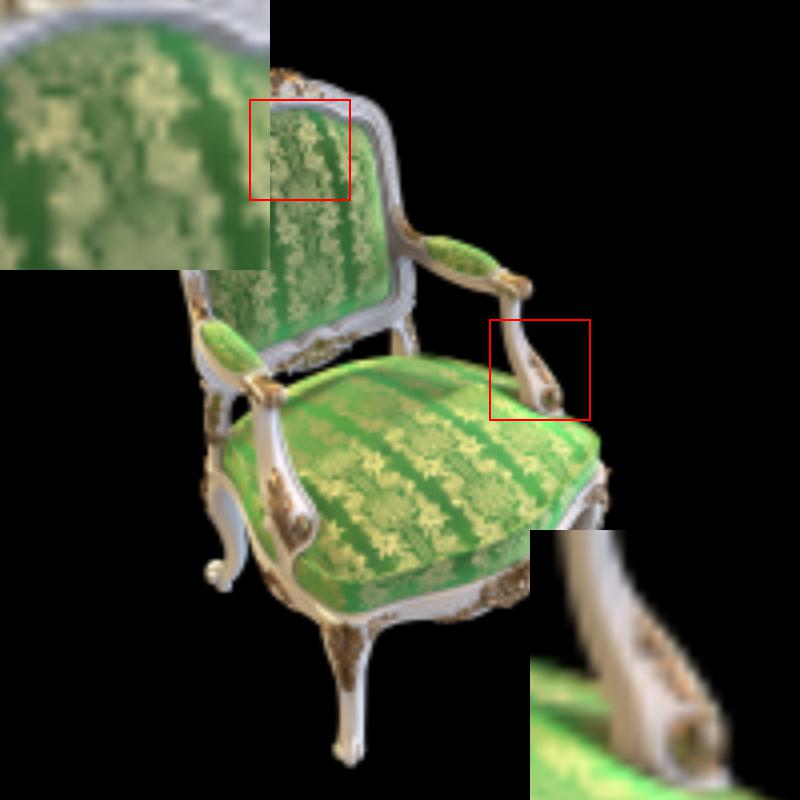} &
    \includegraphics[width=0.248\textwidth,valign=m]{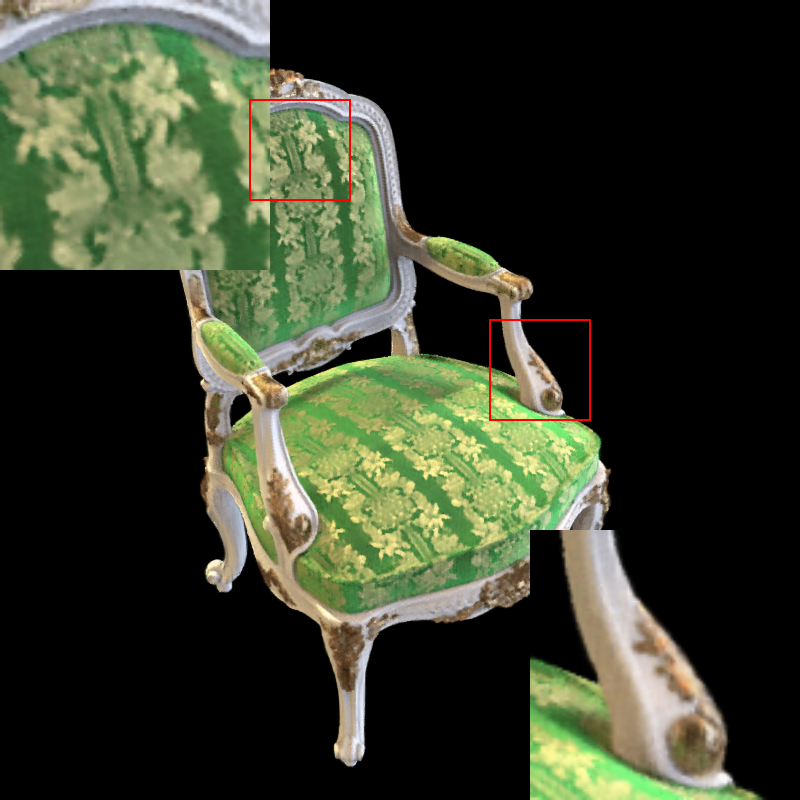} & 
    \includegraphics[width=0.248\textwidth,valign=m]{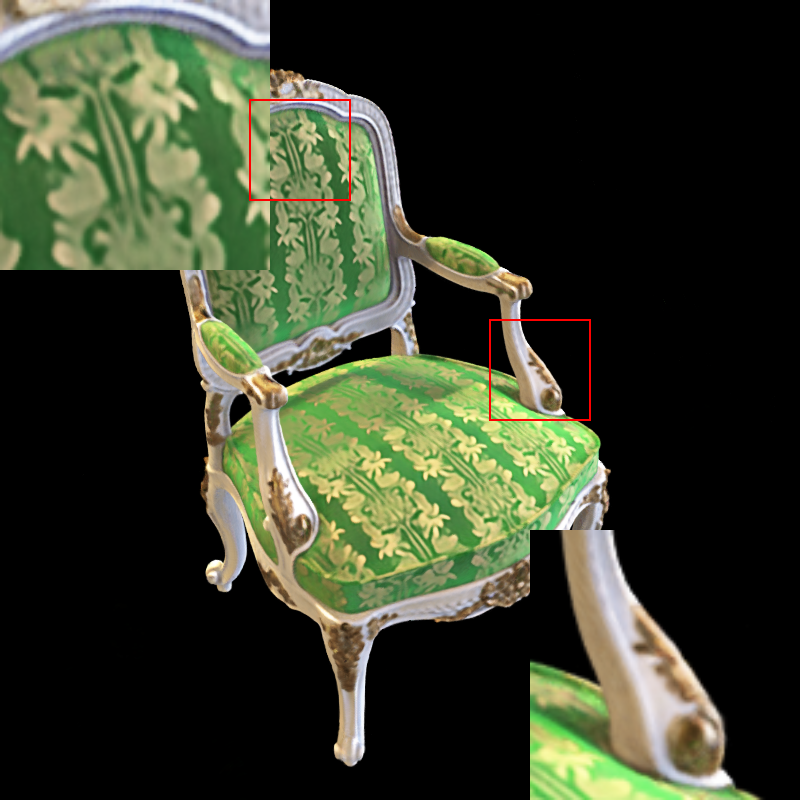} & 
    \includegraphics[width=0.248\textwidth,valign=m]{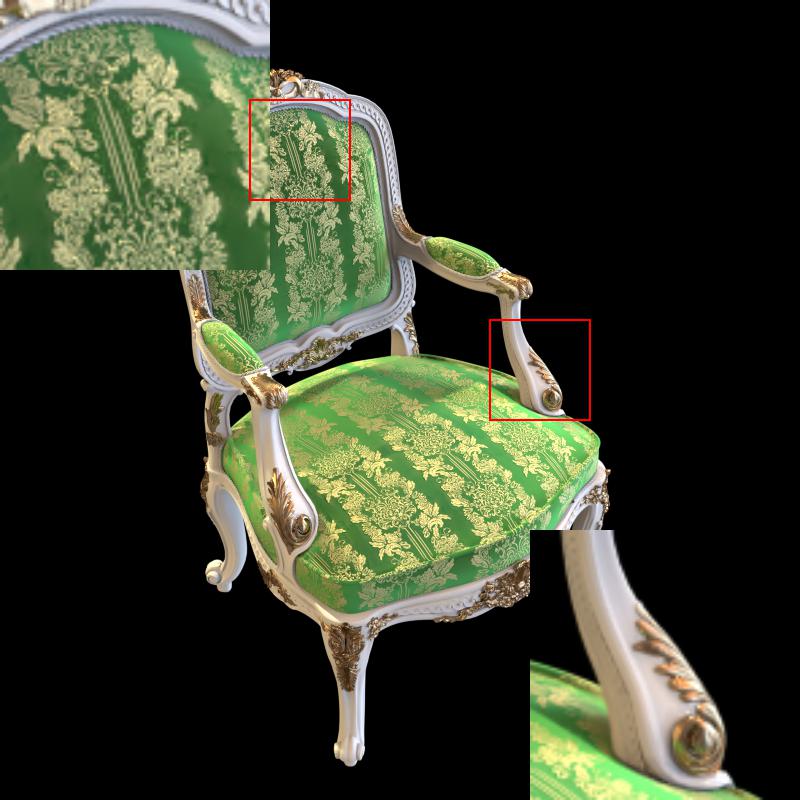} \\

    \includegraphics[width=0.248\textwidth,valign=m]{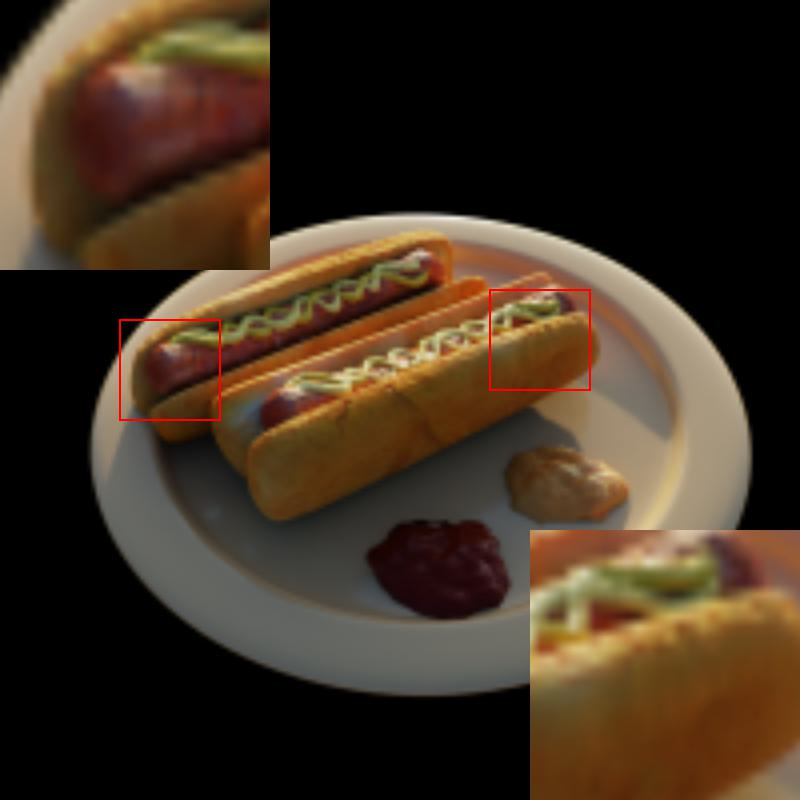} &
    \includegraphics[width=0.248\textwidth,valign=m]{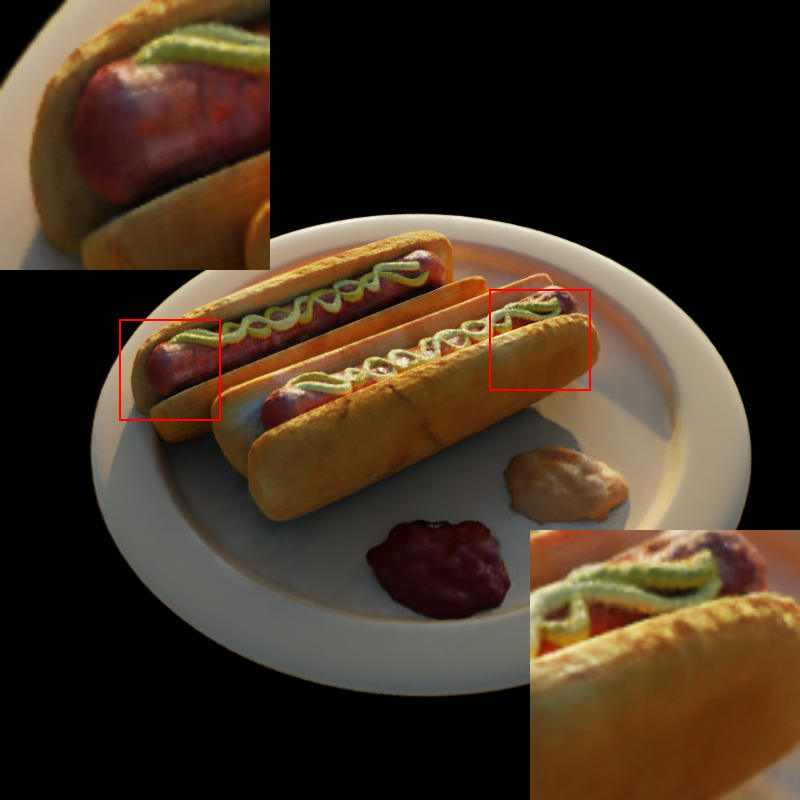} &
    \includegraphics[width=0.248\textwidth,valign=m]{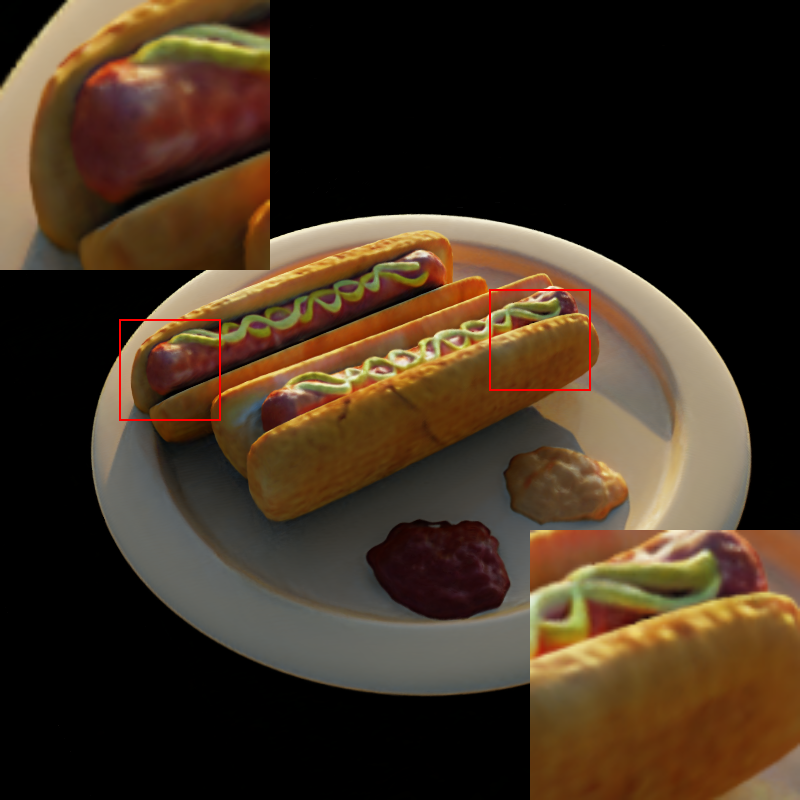} & 
    \includegraphics[width=0.248\textwidth,valign=m]{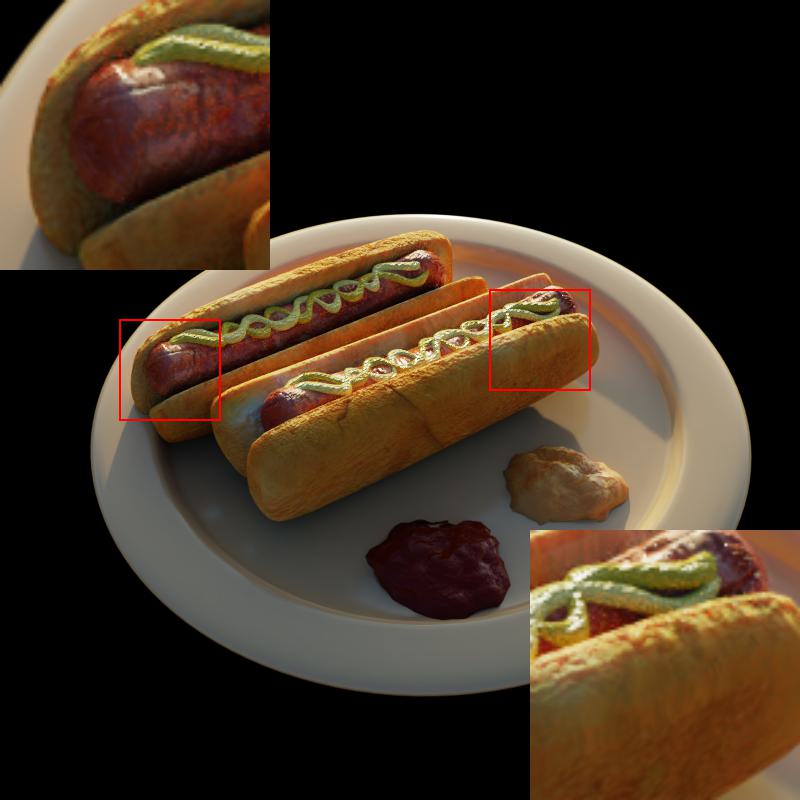} 
    \\
    ~~~~~~ LR image & ~~~~~~~~~ NVSR & ~~ TriNeRFLet SR & ~~HR Ground Truth

    \end{tabular}
    \vspace{-0.2in}
    \caption{\emph{TriNeRFLet SR qualitative results}. Note the visual improvement that is achieved by our approach to NeRF SR compared to LR reconstruction. Moreover, the qualitative results of our method compete with NVSR, which unlike our strategy is a 3D supervised method that we consider as a reference. In the sup. mat. we provide addtional visual examples with more comparisons to other methods.}
    \label{fig:qualitive_results_sr}
    \vspace{-0.1in}
\end{figure}

Results are reported in Table \ref{tab:results_sr}. Note that NVSR requires 3D supervision (paired LR and HR multiview images) and therefore it is considered only as a reference. When compared only to methods that use 2D supervised methods, we perform the best in most of the cases. Also, our method's results are very close to NVSR, and even slightly outperforms it in some cases, which demonstrates the superiority of our method despite the fact that it was not trained using 3D data.
Qualitative examples of our approach appear in Figure \ref{fig:qualitive_results_sr} and sup. mat. 

\noindent \textbf{Impact of multiscale on SR.} To demonstrate the impact of TriNeRFLet multiscale on the SR scheme, we modify it to neutralize the multiscale element. This is done by replacing the low-resolution rendering that uses low-resolution wavelet features (first $L_{LR}$ layers)  with a downscaled HR rendering that uses all wavelet layers (i.e. without multiscale render). In the same manner, INGP is also tested as a representation. Results in Table \ref{tab:sr_ablation} demonstrate the important contribution of the TriNeRFLet multiscale structure to the SR scheme, utilizing its full potential, and that without it the SR scheme performs worse.

\noindent \textbf{Impact of LPIPS loss.} To demonstrate the LPIPS loss impact on SR, we trained the scheme without the LPIPS loss. Results are reported in Table \ref{tab:sr_ablation}, and as noticed, the LPIPS component indeed improves the SR scheme performance.

\begin{table}[t]
    \begin{tabular}{ |p{0.1\textwidth} | p{0.2\textwidth} p{0.21\textwidth} p{0.21\textwidth} p{0.2\textwidth} |}
     \hline
         & \small TriNeRFLetSR  & \small TriNeRFLetSR w/o multiscale & \small TriNeRFLetSR w/o LPIPS & \small Our SR scheme with INGP  \\
     \hline
     
     PSNR$\uparrow$  &  $\textbf{29.49}$  & $28.49$ &  $29.27$ & $28.25$ \\
     \hline

     LPIPS$\downarrow$  & $\textbf{0.051}$  & $0.057$ & $0.06$ & $0.06$ \\
     \hline

     SSIM$\uparrow$  &  $\textbf{0.930}$  & $0.919$ & $0.903$ &  $0.915$  \\
     \hline

    \end{tabular}
    \caption{\textbf{Ablation: Impact of wavelet multiscale and LPIPS on SR.} To show the significance of having the multi-scale structure of the wavelet, we apply our SR scheme without using the multiscale rendering or with INGP instead of TriNeRFLet rendering. We also show the significance of adding the LPIPS loss. }
    \label{tab:sr_ablation}
    \vspace{-0.3in}
\end{table}

\begin{table}[h!]
\centering
\resizebox{\textwidth}{!}{
    \begin{tabular}{ p{0.18\textwidth} | p{0.09\textwidth}  p{0.09\textwidth}  p{0.08\textwidth} |p{0.09\textwidth}  p{0.09\textwidth} p{0.08\textwidth}
    |p{0.09\textwidth}  p{0.09\textwidth} p{0.08\textwidth}  
    }
      & \multicolumn{3}{c|}{Blender} & \multicolumn{3}{c|}{Blender} & \multicolumn{3}{c}{LLFF} \\
      & \multicolumn{3}{c|}{$100 \rightarrow 400$} & \multicolumn{3}{c|}{$200\rightarrow 800$} & \multicolumn{3}{c}{$378 \times 504\rightarrow 1512 \times 2016 $} \\
     Method & PSNR$\uparrow$ & LPIPS$\downarrow$ & SSIM$\uparrow$ & PSNR$\uparrow$ & LPIPS$\downarrow$ & SSIM$\uparrow$ & PSNR$\uparrow$ & LPIPS$\downarrow$ & SSIM$\uparrow$\\
     \hline
     NeRF* & $25.56$ & $0.170$ & $0.881$ & $27.47$ & $0.128$ & $0.900$ &  $24.47$ & $0.388$ & $0.701$\\
     \hline
     NeRF-Bi & $24.74$ & $0.244$ & $0.868$ & $26.67$ & $0.175$ & $0.900$ & $23.90$ & $0.481$ & $0.676$\\
     \hline
     NeRF-Liif & $25.36$ & $0.125$ & $0.885$ & $27.34$ & $0.096$ & $0.912$ & $24.76$ & $0.292$ & $0.723 $\\
     \hline
     NeRF-Swin & $24.85$ & $0.108$ & $0.881$ & $26.78$ & $0.086$ & $0.906$ & $23.26$ & $0.247$ & $0.685$\\
     \hline
     NeRF-SR (SS) &  $\underline{28.07}$ &  $\underline{0.071}$ & $\textbf{0.921}$ & $28.46$ & $0.076$ & $0.921$ &  $\textbf{25.13}$ &  $\underline{0.244}$ & $\underline{0.730}$\\
     \hline
     \hline
    TriNeRFLetSR&  $\textbf{28.55}$ &   $\textbf{0.061}$ & $\underline{0.913}$ &  $\underline{29.49}$ &  $\textbf{0.051}$ & $\underline{0.930}$ &  $\underline{25.00}$  &   $\textbf{0.203}$ & $\textbf{0.771}$\\
     \hline
     \hline
     NVSR \scriptsize (3D supervised) & \multicolumn{3}{c|}{$**$} &  $\textbf{29.53}$  &  $\underline{0.054}$ & $\textbf{0.931}$ & \multicolumn{3}{c}{$**$}\\
     \hline
    \end{tabular}
}
    \caption{\emph{Super Resolution $\times 4$ Results}. NVSR is trained with multiview (3D) supervision and therefore we consider it as reference. \textbf{Bold} is best, \underline{underline} is second. **For $100 \rightarrow 400$ NVSR reports results of only 4 shapes in the Blender dataset. In this case, we outperform it with PSNR of $29.18$dB for our method compared to $28.5$dB of NVSR. For LLFF, NVSR did not report any results on this resolution or LLFF average.}
    \label{tab:results_sr}
     \vspace{-0.2in}
\end{table}

\noindent \textbf{LLFF.} We turn to demonstrate the TriNeRFLet-based SR scheme on the LLFF dataset \cite{mildenhall2019local}, which contains 8 real-world captured scenes. As in the Blender SR case, we compare our method with the same methods but for input resolution $378 \times 504$ and $\times 4$ upscaling. 
Following \cite{wang2022nerf_sr}, we trained TriNeRFLet using the low-resolution version of all images and then tested the SR results also on all the images.
Table \ref{tab:results_sr} reports the results. 
Our method is almost as good as \cite{wang2022nerf_sr} in PSNR and outperforms it in the LPIPS and SSIM metrics by a margin. Our LPIPS and SSIM performance indicates that our method manages to create a scene with better visual details and that is perceptually more similar to the ground-truth high-resolution scene. 

\section{Conclusions}
\label{sec:conclusions}

This paper introduced a new NeRF structure that relies on a multiscale wavelet representation. This structure improved the performance of Triplane, which lagged behind NeRF state-of-the-art methods, achieving competitive results. Having the multiscale structure enabled us to implement a new diffusion-guided SR for NeRF. We believe that the concept of learning the features in the multiscale wavelet domain instead of the original one has the potential to improve other vision-related applications.

Despite the advantages of our approach, it also has some limitations, mainly due to the runtime overhead it adds to training time. This can be partially mitigated by using its light-weight versions and the coarse to fine training, which reduce training time. We defer further run-time improvements to future work.

\textbf{Acknowledgments} This work was supported
in part by KLA grant.

{
    \small
    \bibliographystyle{splncs04}
    \bibliography{main}
}

\appendix

\clearpage
\title{TriNeRFLet: A Wavelet Based Triplane NeRF Representation \\ Supplementary Materials} 
\titlerunning{TriNeRFLet Supplementary Materials}
\author{}
\institute{}
\maketitle

\section{Ablation}
\label{sec:abl}
\begin{figure}[h]
    \centering
    \begin{subfigure}[t]{0.9\textwidth} 
        \centering
        \includegraphics[width=\textwidth]{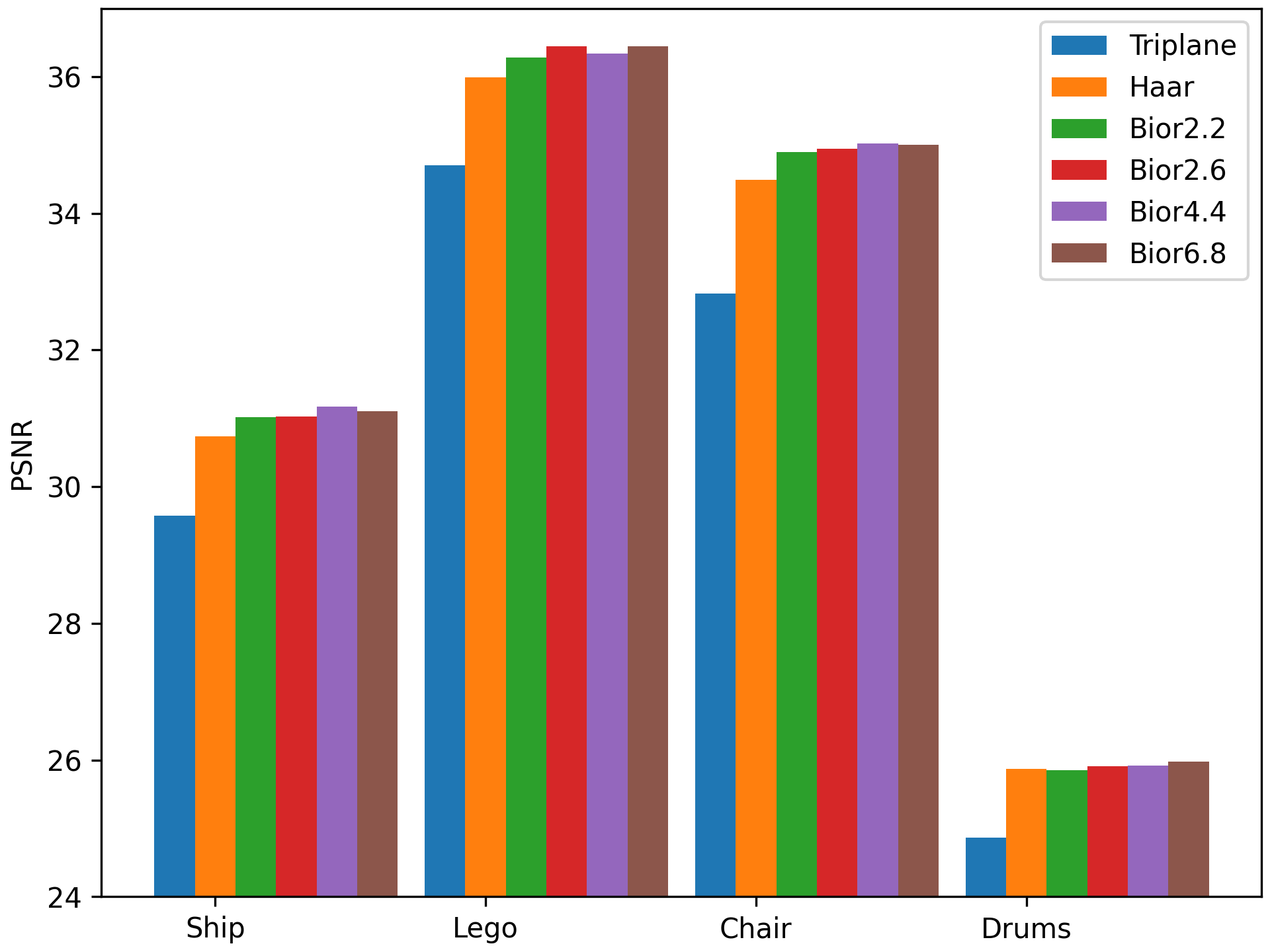}
    \end{subfigure}%
    \caption{Performance comparison between different wavelet filters and no wavelet at all (Triplane).}
    \label{figure:ablation_wavelet}
\end{figure}

\textbf{Wavelet type}. For all the experiments we conducted in the paper, we applied the Biorthogonal 6.8 (Bior6.8) wavelet. We present here an ablation experiment that checks the impact of other wavelet filters. We compare the reconstruction performance of 4 scenes from the Blender dataset. The setups we tested are vanilla Triplane (no wavelet), Haar, Biorthogonal 2.2 (Bior2.2), Biorthogonal 2.6 (Bior2.6), Biorthogonal 4.4 (Bior4.4) and Biorthogonal 6.8 (Bior6.8). The results are presented in Figure \ref{figure:ablation_wavelet}. We use the TriNeRFLet Base Light setting. Vanilla Triplane has the same size and structure as this setting. Note that all wavelets achieve a significant performance advantage over vanilla Triplane (as shown also in the paper). 
Generally, higher order wavelets provide a better representation of smooth functions \cite{Mallat2008Wavelet}. Thus, it is not surprising that Bior6.8, Bior4.4 and Bior 2.6 achieve better performance than Bior 2.2 and Haar. Note though that in terms of training time, training with Haar wavelet is faster by up to $30\%$ compared to the other wavelets as Haar complexity is $O(N)$ and the complexity of applying the other wavelets is $O(N \log(N))$.

\noindent \textbf{Coarse To Fine}. As mentioned in the paper, coarse to fine (c2f) accelerates TriNeRFLet framework and is presented as a mitigation for the wavelet inverse transform overhead. To further check the additional contributions of c2f, we conduct the Blender reconstruction experiment (Base Light version) but without c2f. Results are presented in Table \ref{tab:wavelet_reg_ablation}, and as noticed, c2f indeed brings a contribution in the reconstruction quality, alongside the contribution in training time.

\noindent \textbf{Wavelet Regularization}. To check the importance of the wavelet regularization loss, we extend the c2f experiment above to be without wavelet regularization. The results in Table \ref{tab:wavelet_reg_ablation} illustrate the important contribution of wavelet regularization and that it is an essential component of TriNeRFLet performance.

\section{Technical Details}
\label{sec:tech_details}
\textbf{3D Reconstruction}. Table \ref{tab:versions} contains the parameters for the reconstruction versions.

\noindent \textbf{Blender Super-Resolution.} In both Blender settings ($100 \rightarrow 400 \And 200 \rightarrow 800$), we use $C=16$ (channels), $s_{LR}=6000$ (LR only steps), $s=16000$ (total steps), $s_{refresh}=500$ (refresh steps), $T_{min}=0.02$ and $T_{max}$ is initialized with $0.98$ and decreases linearly to reach $0.25$ at the end. We use \cite{threestudio2023} as a base for implementing the super-resolution scheme, as it is more suitable for this task. We run the SR method on a single A$6000$ RTX GPU, and it runs for almost $7$ hours. 

\noindent \textbf{LLFF Super-Resolution.} We use similar settings to Blender $200\rightarrow 800$ experiment except for, $C=32$ (channels), $s=20000$ (total steps), $s_{refresh}=250$ (refresh steps). Like \cite{mildenhall2021nerf}, we also operate in NDC space coordinates, as it fits this type of scenes better. Similar to the Blender experiment, we ran LLFF SR method on a single A6000 RTX GPU, where each scene runs almost for $12$ hours.

\begin{table}[h!]
\centering
    \begin{tabular}{ |p{0.07\textwidth} | p{0.06\textwidth} p{0.03\textwidth} p{0.07\textwidth} p{0.08\textwidth} p{0.04\textwidth} p{0.04\textwidth} p{0.04\textwidth} p{0.08\textwidth} p{0.06\textwidth} p{0.06\textwidth} p{0.14\textwidth}| }
     \hline
       & {\small $N_{LL}$} & \small $L$ & $N_{base}$ & \small $N_{final}$ & \small$C$ & \small $\gamma$ & \small $W$ & \small $D_{dens}$ & \small $D_{col}$ & \small train steps & \small trainable parameters \\
     \hline
     
     Small & $64$ & $4$ &$512$ & $1024$ & $16$ & $0.2$ & $64$ & $1$ & $2$ & $6k$ & $17M$\\
     \hline

     Base Light & $64$ & $5$ &$512$ & $2048$ & $32$ & $0.4$ & $64$ & $1$ & $2$ & $10k$ & $134M$\\
     \hline

     Base & $64$ & $5$ &$512$ & $2048$ & $32$ & $0.4$ & $64$ & $1$ & $2$ & $43k$ & $134M$\\
     \hline

     Large & $64$ & $5$ &$512$& $2048$ & $48$ & $0.6$ & $128$ & $1$ & $2$ & $83k$ & $201M$\\
     \hline
     
    \end{tabular}
    \caption{The configurations used in the different TriNeRFLet versions}
    \label{tab:versions}
\end{table}

\begin{table}[t!]
\centering
    \begin{tabular}{ |p{0.13\textwidth} | p{0.07\textwidth} p{0.07\textwidth} p{0.07\textwidth} p{0.11\textwidth} p{0.07\textwidth} p{0.09\textwidth} p{0.07\textwidth} p{0.1\textwidth} | p{0.06\textwidth} |  p{0.07\textwidth} |}
     \hline
       & {\small Mic} & \small Chair & \small Ship & \small Materials & \small Lego & \small Drums & \small Ficus & \small Hotdog & \small Avg. &Train Time \\
     \hline

     Base Light& $35.77$ &$35.00$& $31.10$ & $29.35$ & $36.44$ & $25.98$ & $33.96$ & $36.93$ & $33.07$ & $1.5$ hours \\
     \hline

    Base Light w/o c2f & $35.52$ & $35.08$& $31.11$ & $29.14$ & $36.48$ & $25.76$ & $33.46$ & $36.88$ & $32.93$ & $2$ hours \\
     \hline

     Base Light w/o c2f $\And$ wavelet reg. & $34.14$ & $33.33$& $30$ & $28.17$ & $35.02$ & $25.01$ & $30.76$ & $35.95$ & $31.55$ & $2$ hours \\
     \hline

     Triplane& $33.85$ &$32.83$& $29.58$ & $28.15$ & $34.70$ & $24.86$ & $30.35$ & $35.80$ & $31.26$ & $2$ hours \\ 
     \hline

    \end{tabular}
    \caption{Coarse to fine (c2f) and wavelet regularization importance ablation. We compare Base Light version with the versions without c2f and without c2f and wavelet regularization. The results indicate the important contribution of the c2f and wavelet regularization.}
    \label{tab:wavelet_reg_ablation}
\end{table}

\section{Detailed Results}
\label{sec:add_res}
Tables \ref{tab:blender100_sr},\ref{tab:blender200_sr} and \ref{tab:LLFF_sr} contain the detailed results of the SR experiments described in the paper.

\begin{table}[h!]
\centering
    \begin{tabular}{ |p{0.13\textwidth} | p{0.07\textwidth} p{0.07\textwidth} p{0.07\textwidth} p{0.11\textwidth} p{0.07\textwidth} p{0.09\textwidth} p{0.07\textwidth} p{0.1\textwidth} | p{0.06\textwidth} | }
     \hline
       & {\small Mic} & \small Chair & \small Ship & \small Materials & \small Lego & \small Drums & \small Ficus & \small Hotdog & \small Avg. \\
     \hline

     PSNR & $31.54$ &$29.4$& $28.11$ & $27.9$ & $27.65$ & $24.18$ & $27.7$  & $31.92$ & $28.55$ \\
     \hline

    LPIPS & $0.05$ &$0.051$& $0.118$ & $0.048$ & $0.073$ & $0.067$ & $0.044$  & $0.039$  & $0.061$ \\
     \hline

     SSIM & $0.914$ &$0.946$& $0.844$ & $0.921$ & $0.914$ & $0.896$ & $0.936$  & $0.939$ & $0.913$ \\
     \hline

    \end{tabular}
    \caption{Blender $100 \rightarrow 400$ detailed results.}
    \label{tab:blender100_sr}
\end{table}

\begin{table}[h!]
\centering
    \begin{tabular}{ |p{0.13\textwidth} | p{0.07\textwidth} p{0.07\textwidth} p{0.07\textwidth} p{0.11\textwidth} p{0.07\textwidth} p{0.09\textwidth} p{0.07\textwidth} p{0.1\textwidth} | p{0.06\textwidth} | }
     \hline
       & {\small Mic} & \small Chair & \small Ship & \small Materials & \small Lego & \small Drums & \small Ficus & \small Hotdog & \small Avg. \\
     \hline

     PSNR & $31.11$ &$29.86$ & $28.14$ & $28.02$ & $30.46$ & $24.17$ & $30.6$  & $33.58$ & $29.49$ \\
     \hline

    LPIPS & $0.032$ &$0.059$ & $0.114$ & $0.046$ & $0.032$ & $0.071$ & $0.021$  & $0.034$ & $0.051$ \\
     \hline

     SSIM & $0.966$ &$0.944$ & $0.844$ & $0.93$ & $0.922$ & $0.913$ & $0.96$  & $0.96$ & $0.93$ \\
     \hline

    \end{tabular}
    \caption{Blender $200 \rightarrow 800$ detailed results.}
    \label{tab:blender200_sr}
\end{table}

\begin{table}[h!]
\centering
    \begin{tabular}{ |p{0.13\textwidth} | p{0.08\textwidth} p{0.07\textwidth} p{0.08\textwidth} p{0.11\textwidth} p{0.07\textwidth} p{0.09\textwidth} p{0.07\textwidth} p{0.1\textwidth} | p{0.06\textwidth} | }
     \hline
       & {\small Flower} & \small Fern & \small Leaves & \small Fortress & \small Horns & \small Orchids & \small Room & \small Trex & \small Avg. \\
     \hline

     PSNR & $26.6$ &$24$ & $20.01$ & $28.88$ & $24.93$ & $21.38$ & $28.56$  & $25.57$ & $25.00$ \\
     \hline

     LPIPS & $0.214$ &$0.205$ & $0.239$ & $0.175$ & $0.26$ & $0.244$ & $0.125$  & $0.163$ & $0.203$ \\
     \hline

     SSIM & $0.808$ &$0.766$ & $0.639$ & $0.832$ & $0.737$ & $0.668$ & $0.901$  & $0.819$ & $0.771$ \\
     \hline

    \end{tabular}
    \caption{LLFF $378 \times 504\rightarrow 1512 \times 2016 $ detailed results.}
    \label{tab:LLFF_sr}
\end{table}

\section{Qualitative Results}
\label{sec:qua_sec}

\begin{figure}[h!]
    \resizebox{\columnwidth}{!}{\begin{tabular}{   p{0.2\textwidth}  p{0.2\textwidth}  p{0.2\textwidth} p{0.2\textwidth}}
    
       \multicolumn{1}{c}{Triplane} & \multicolumn{1}{c}{TriNeRFLet} & 
       \multicolumn{1}{c}{INGP} & \multicolumn{1}{c}{Ground-Truth}\\

    \includegraphics[width=0.21\textwidth,valign=m]{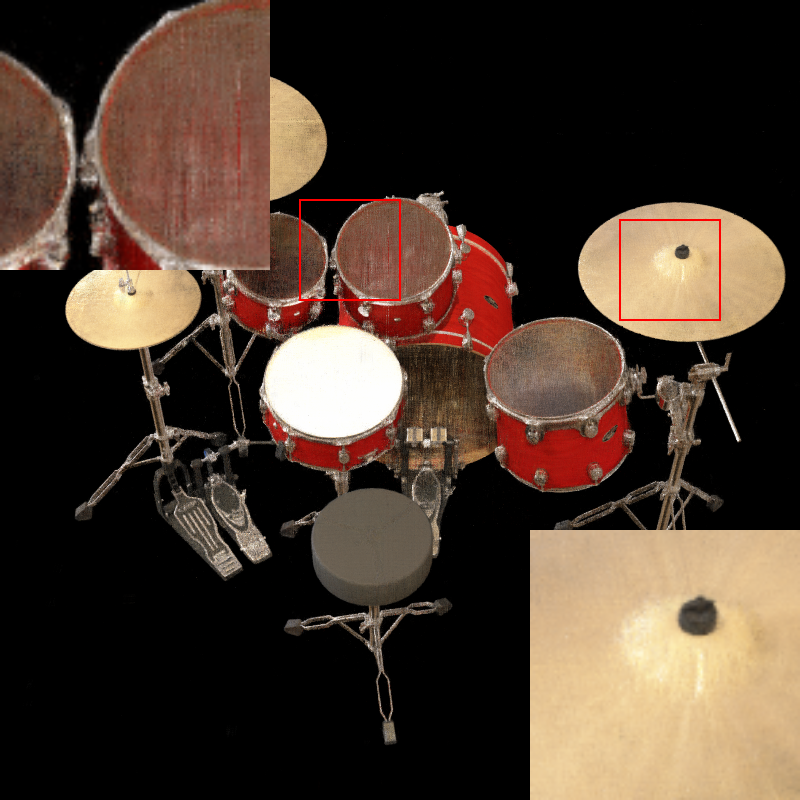} & 
    \includegraphics[width=0.21\textwidth,valign=m]{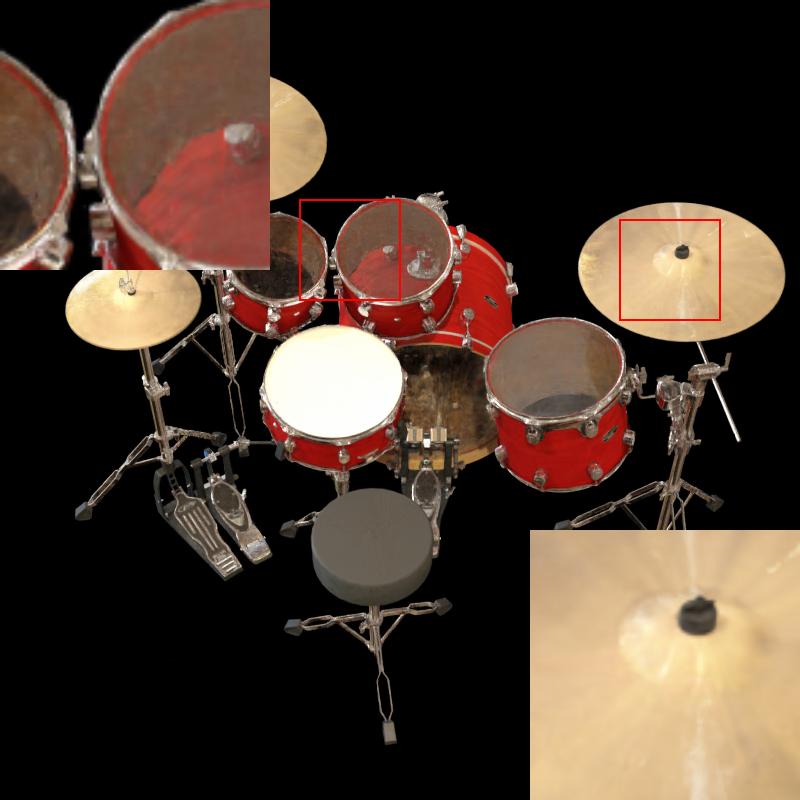} &
    \includegraphics[width=0.21\textwidth,valign=m]{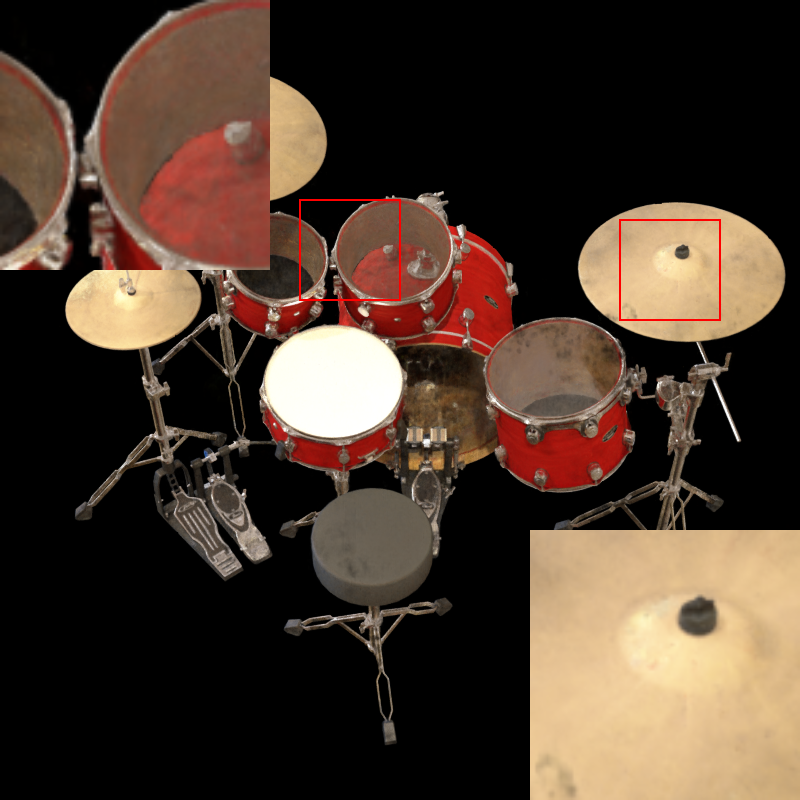} &
    \includegraphics[width=0.21\textwidth,valign=m]{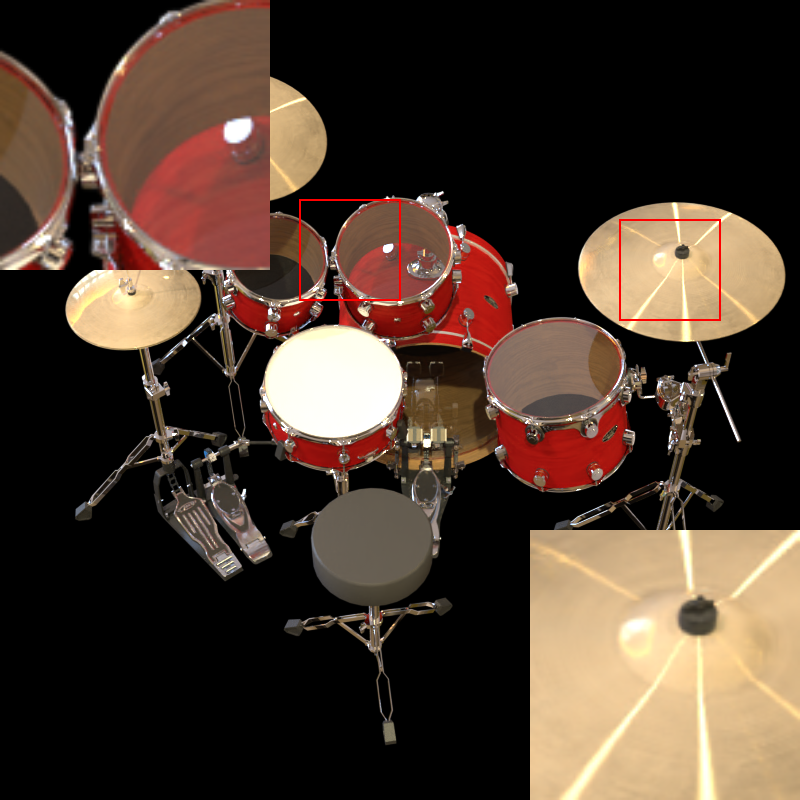} \\

    \includegraphics[width=0.21\textwidth,valign=m]{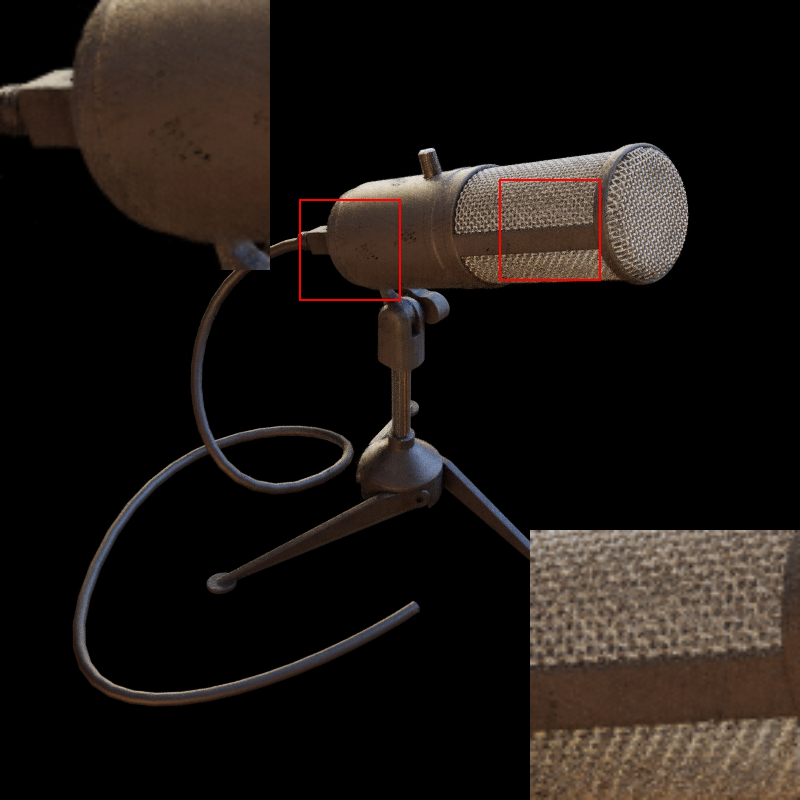} & 
    \includegraphics[width=0.21\textwidth,valign=m]{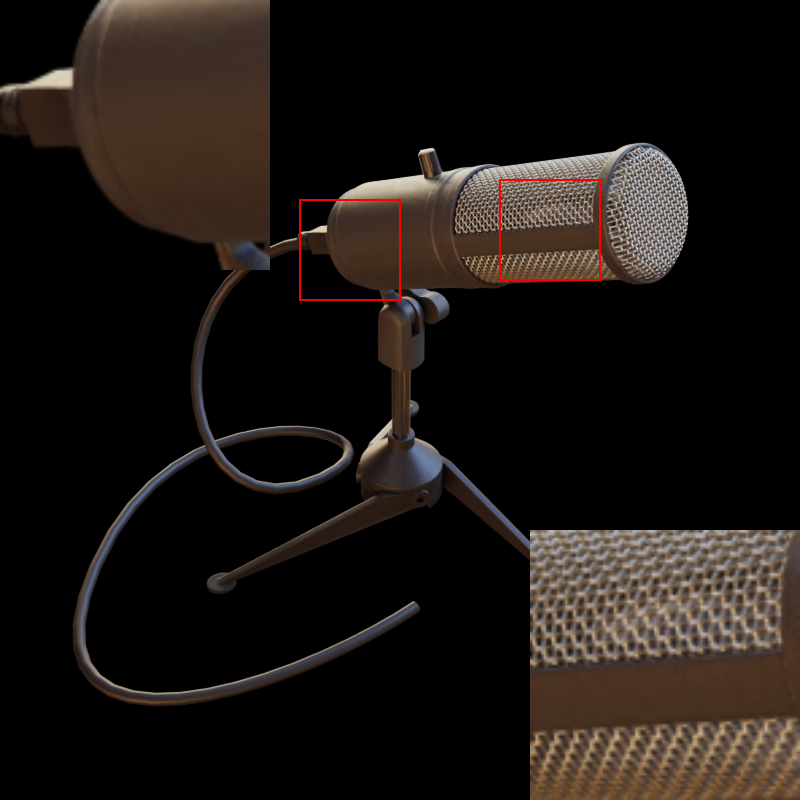} &
    \includegraphics[width=0.21\textwidth,valign=m]{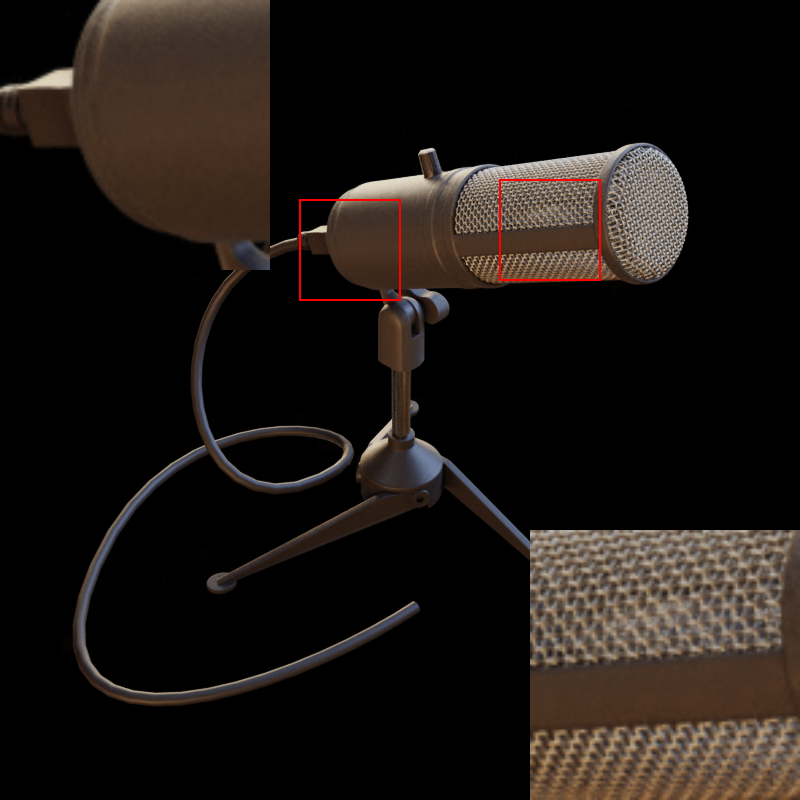} &
    \includegraphics[width=0.21\textwidth,valign=m]{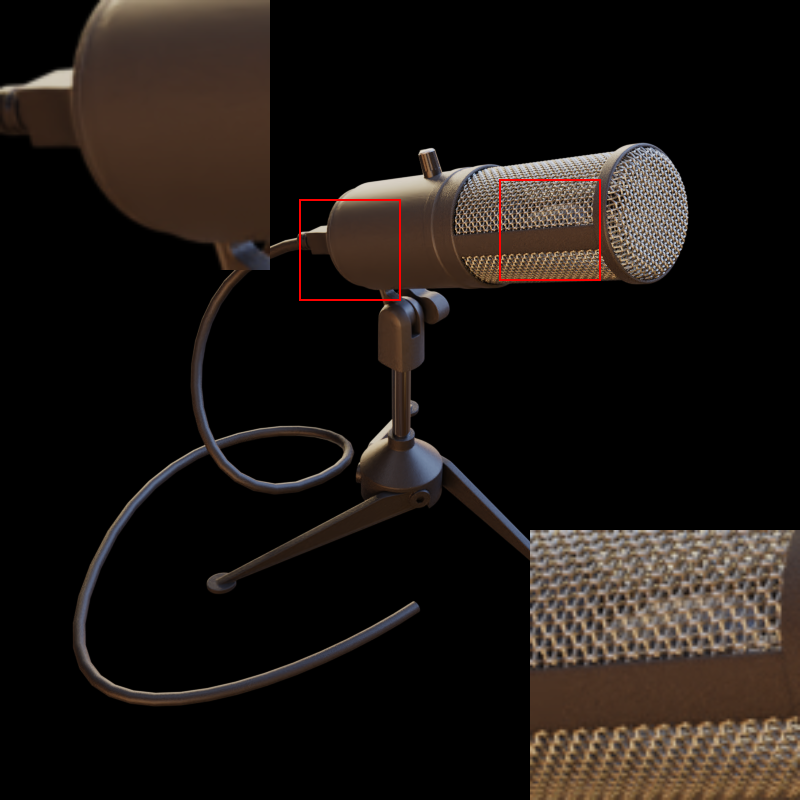} \\

    \includegraphics[width=0.21\textwidth,valign=m]{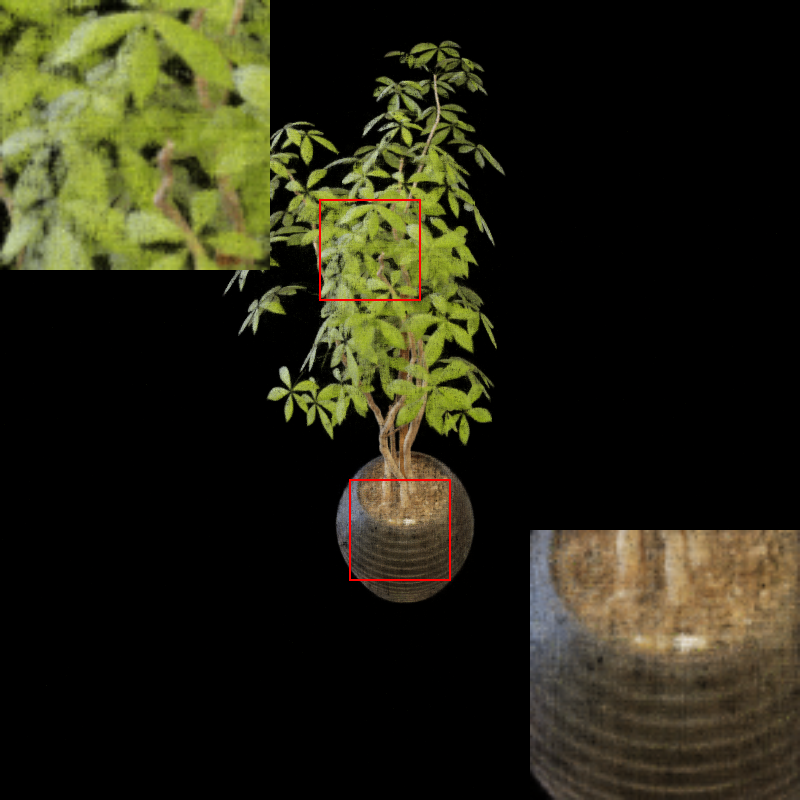} & 
    \includegraphics[width=0.21\textwidth,valign=m]{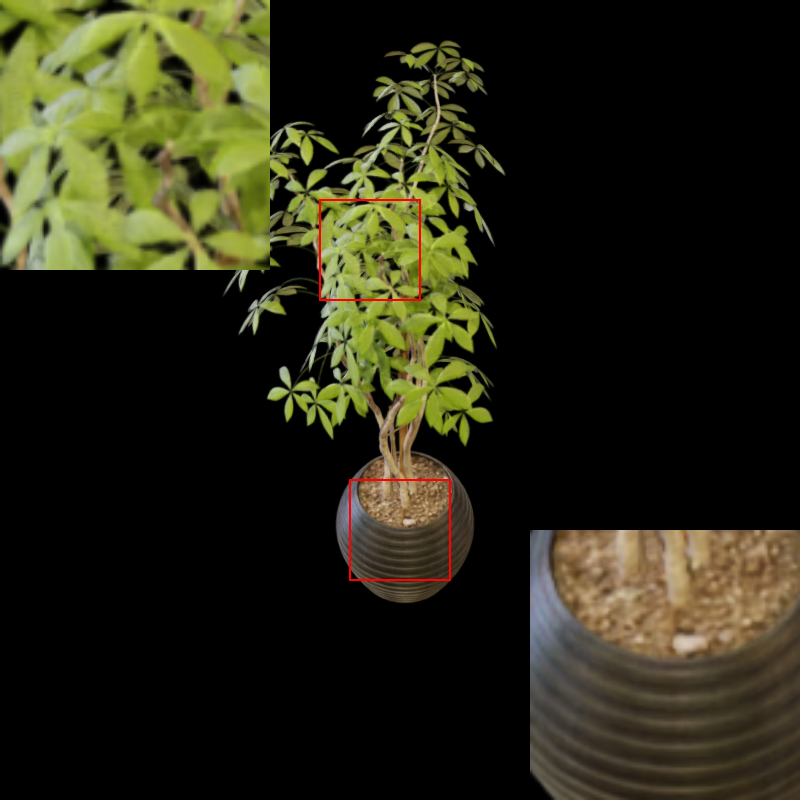} &
    \includegraphics[width=0.21\textwidth,valign=m]{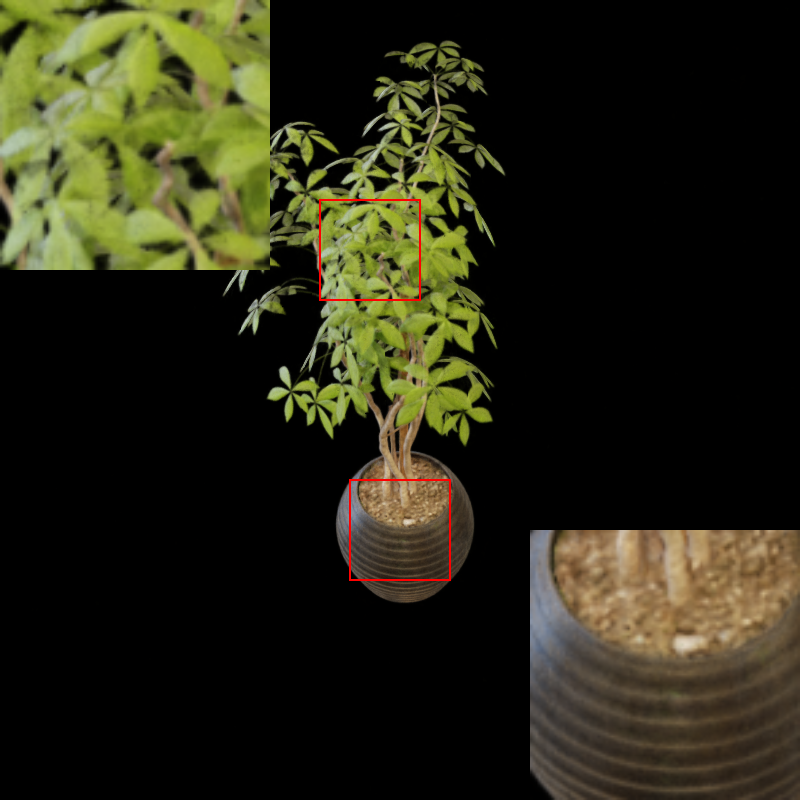} &
    \includegraphics[width=0.21\textwidth,valign=m]{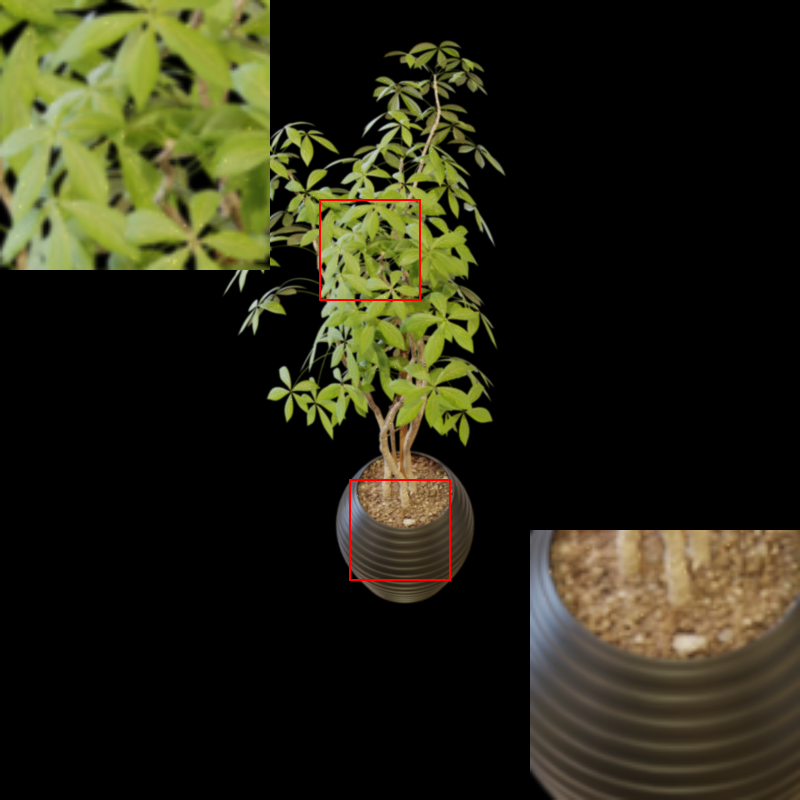} \\

    \end{tabular}}
    \caption{\emph{NeRF reconstruction qualitative results}. Notice the improvement in reconstruction quality of TriNeRFLet compared to Triplane. More visual results appear in the project page.}
    \label{fig:qualitive_results_rec}
\end{figure}

\begin{figure}[h]
    \resizebox{\columnwidth}{!}{\begin{tabular}{   p{0.2\textwidth}  p{0.2\textwidth}  p{0.2\textwidth} p{0.2\textwidth}}
    
       \multicolumn{1}{c}{LR} & \multicolumn{1}{c}{NeRF-SR} & 
       \multicolumn{1}{c}{TriNeRFLet} & \multicolumn{1}{c}{HR}\\

    \includegraphics[width=0.21\textwidth,valign=m]{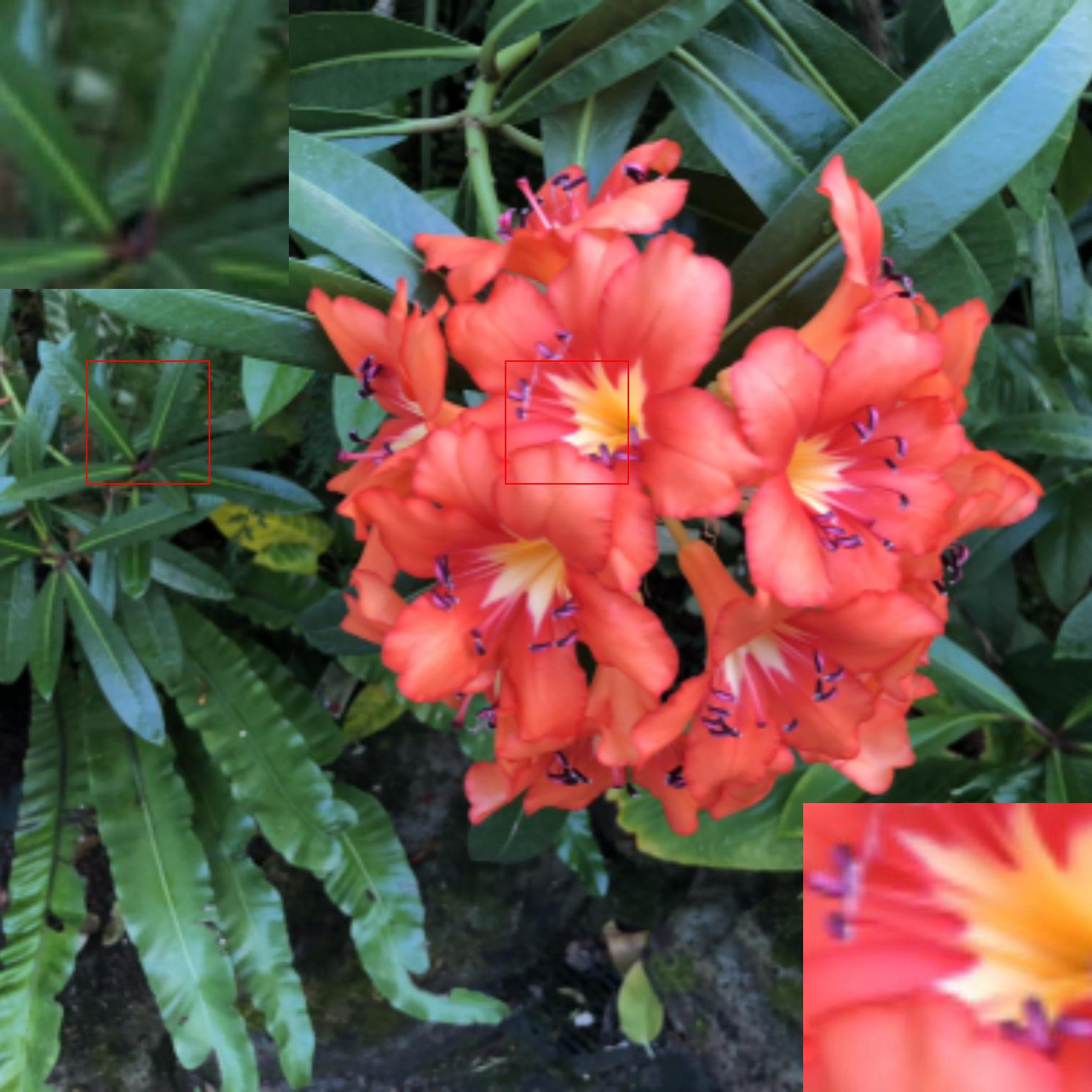} & 
    \includegraphics[width=0.21\textwidth,valign=m]{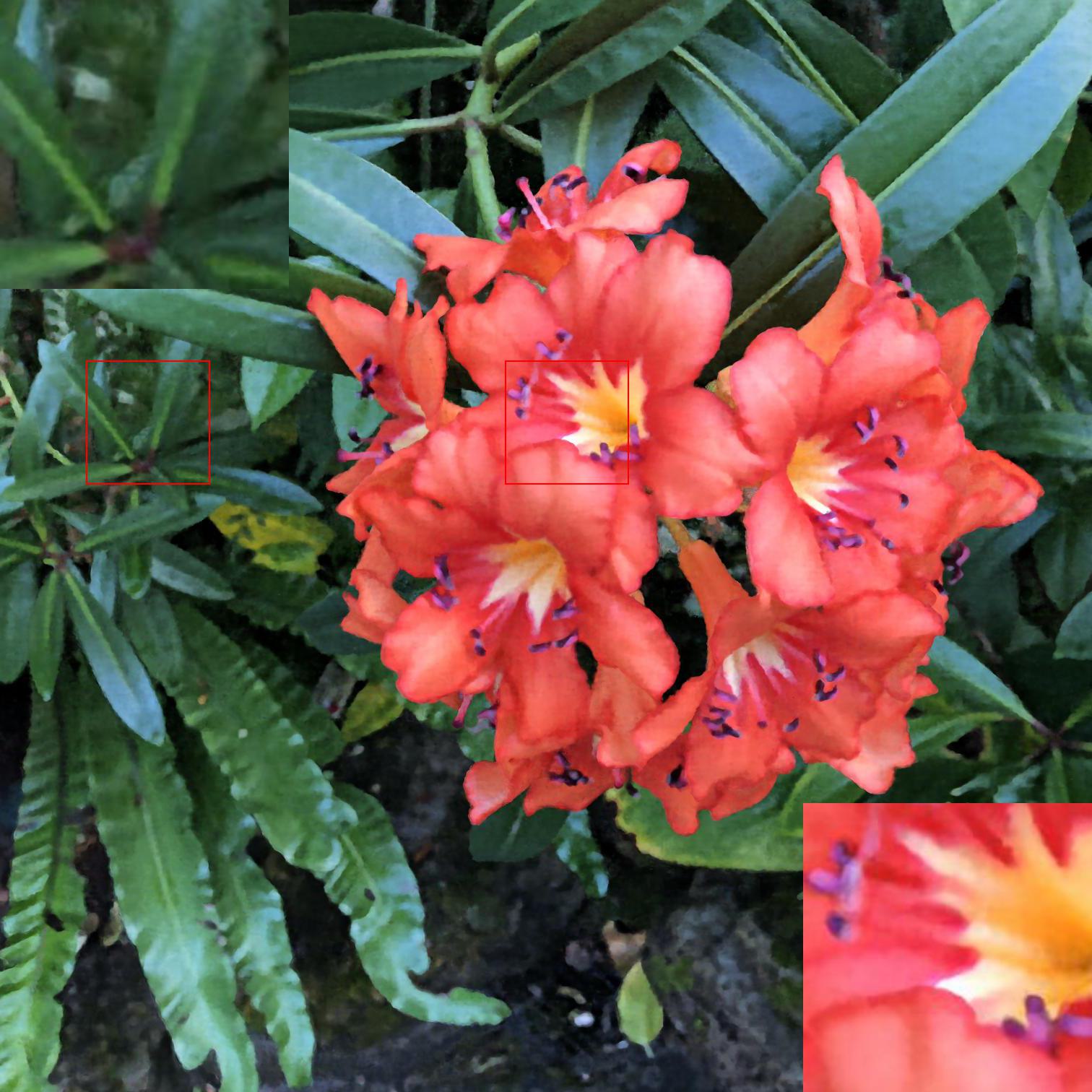} &
    \includegraphics[width=0.21\textwidth,valign=m]{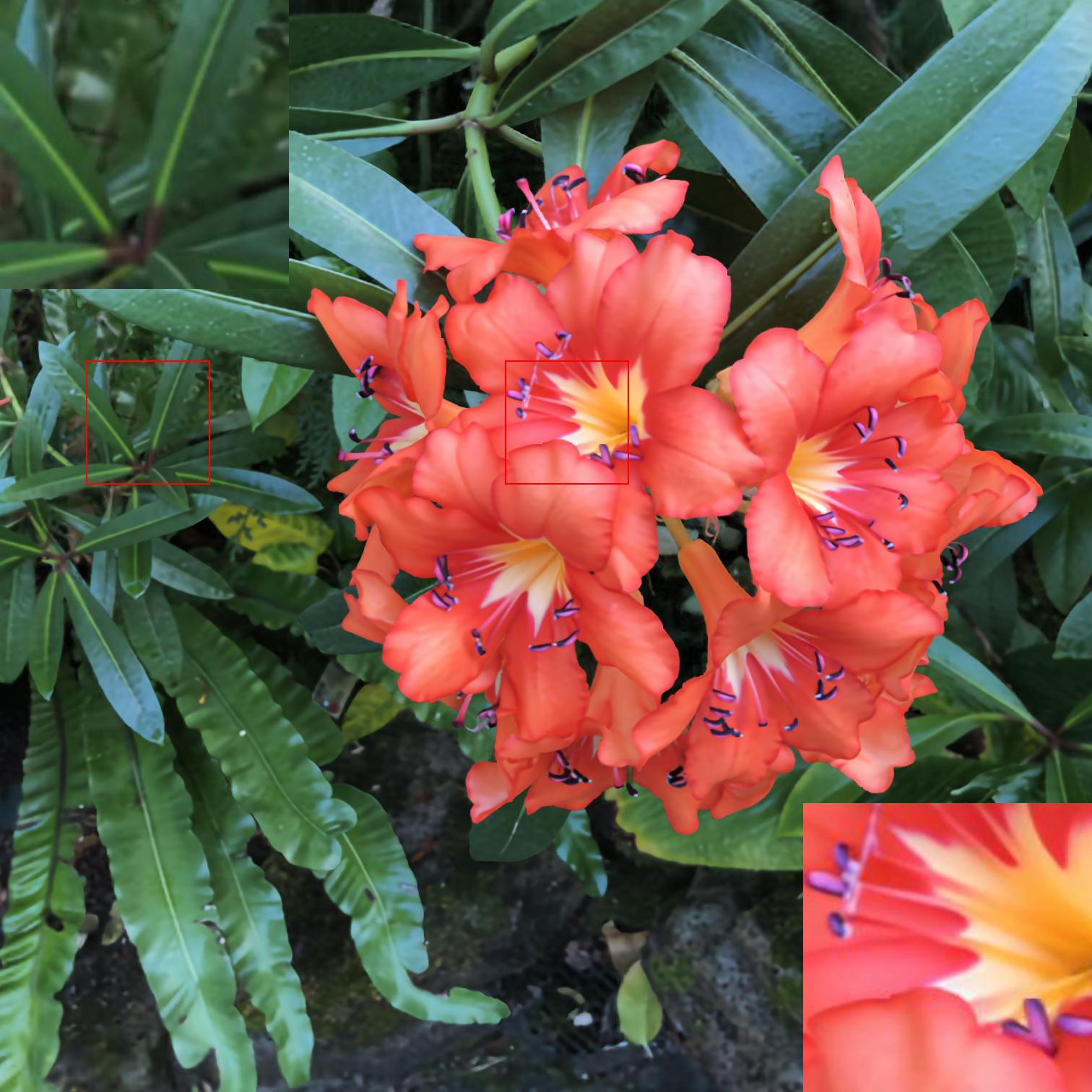} &
    \includegraphics[width=0.21\textwidth,valign=m]{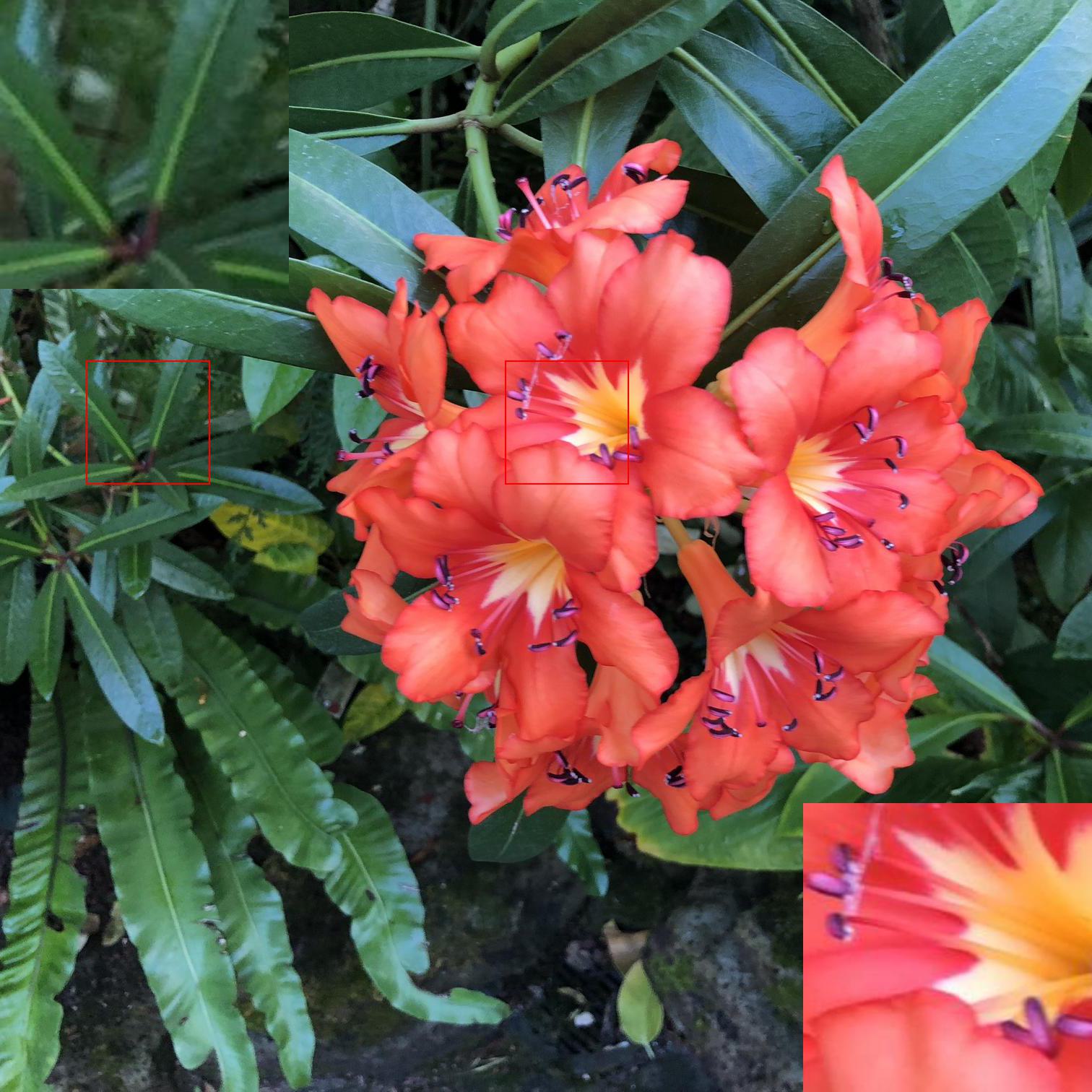} \\

    \includegraphics[width=0.21\textwidth,valign=m]{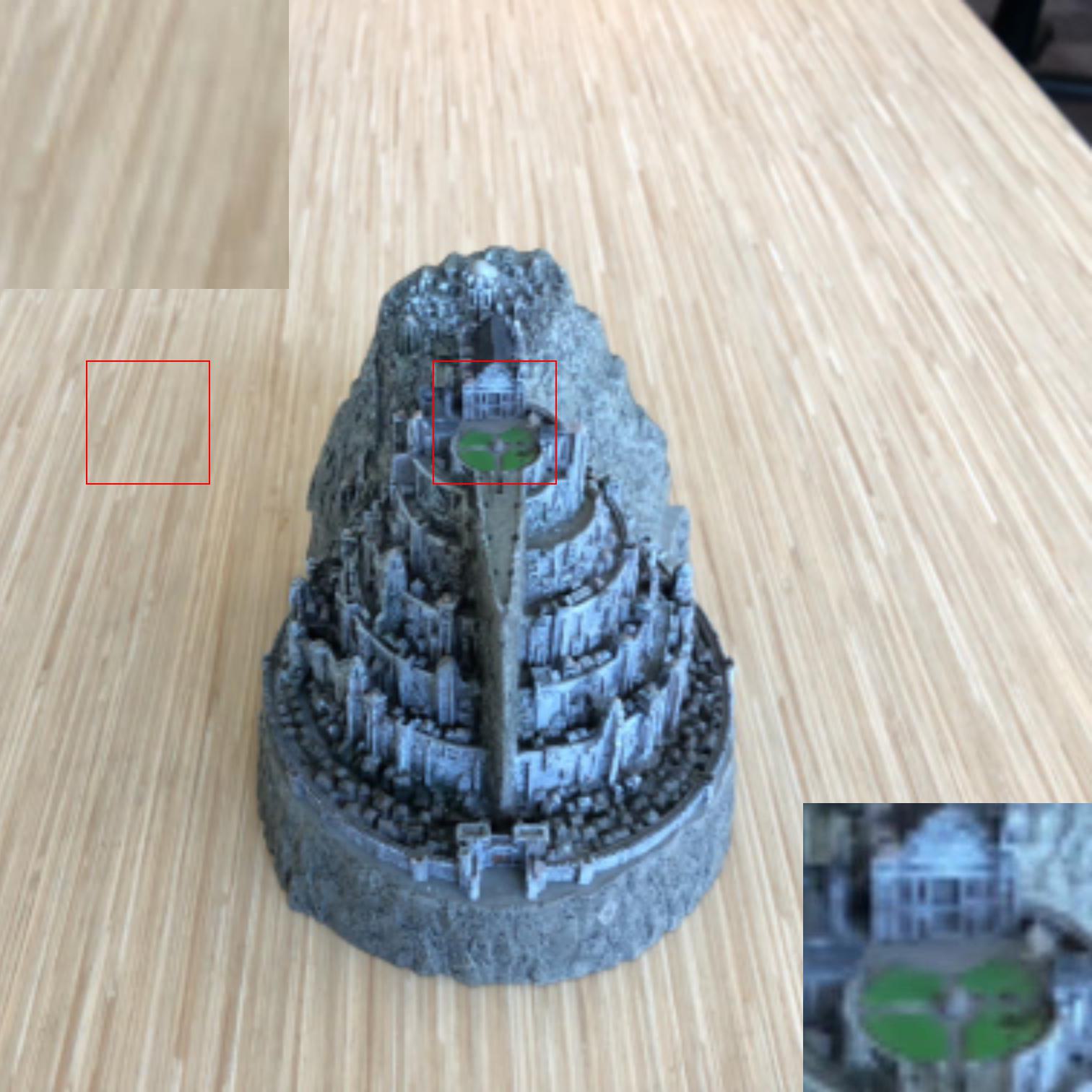} & 
    \includegraphics[width=0.21\textwidth,valign=m]{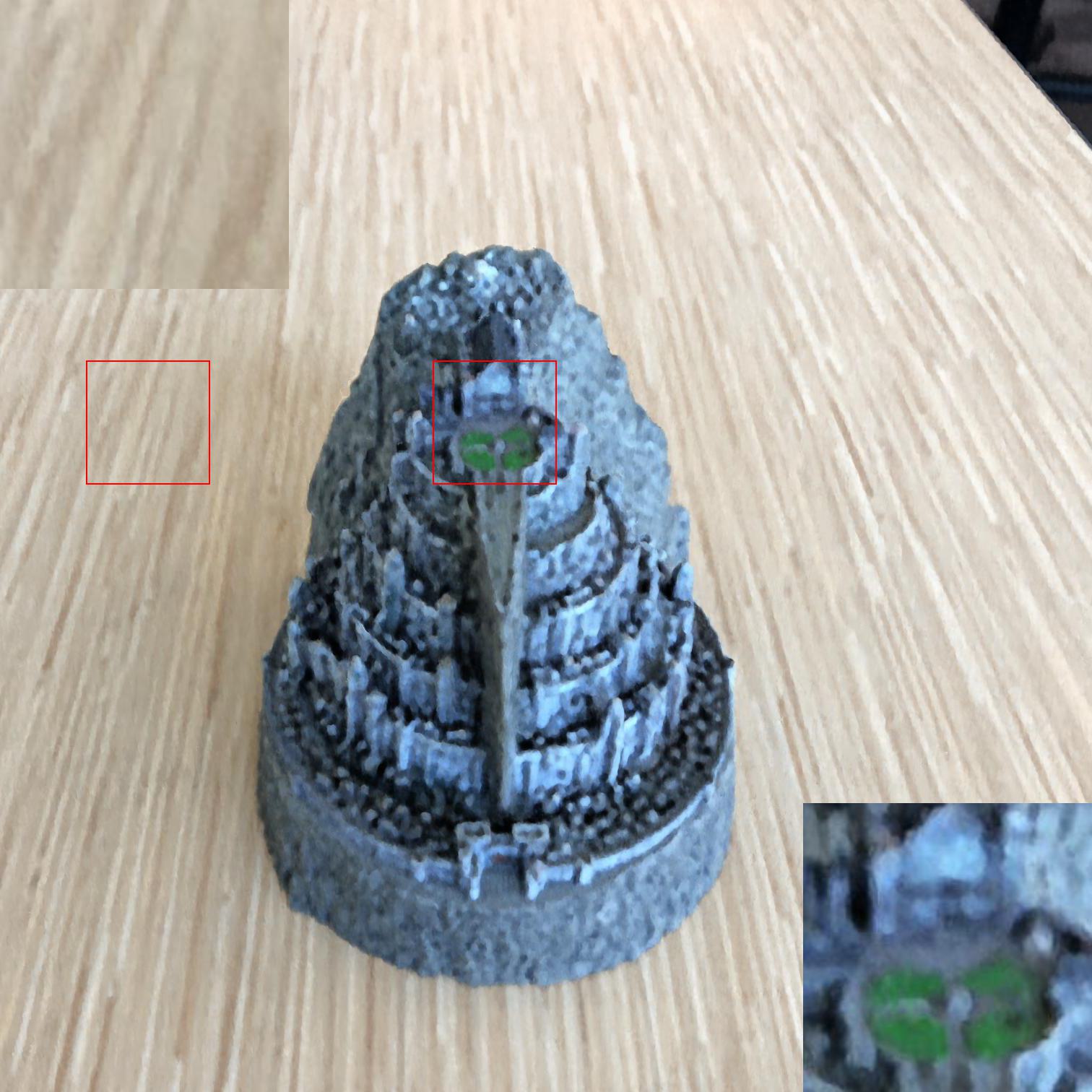} &
    \includegraphics[width=0.21\textwidth,valign=m]{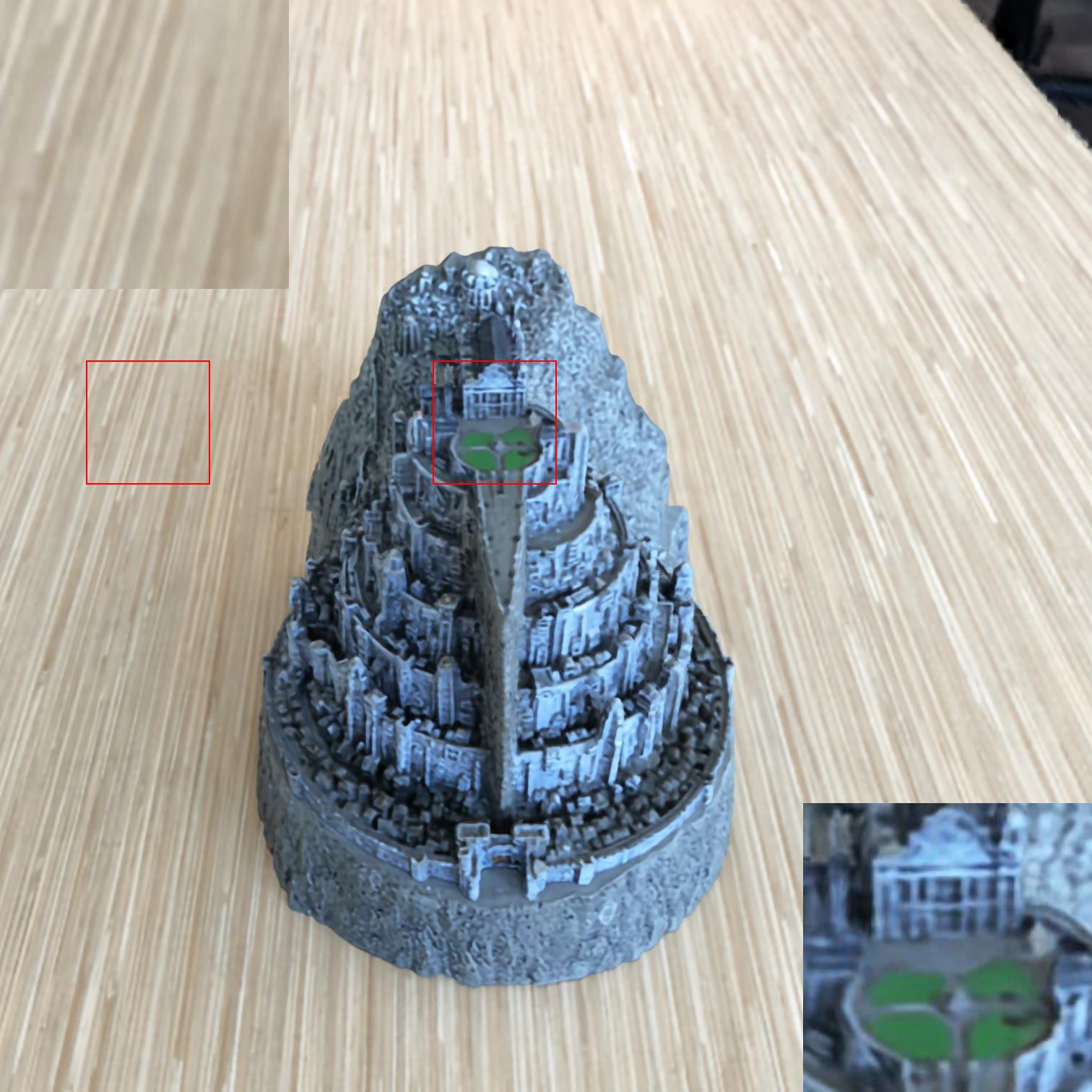} &
    \includegraphics[width=0.21\textwidth,valign=m]{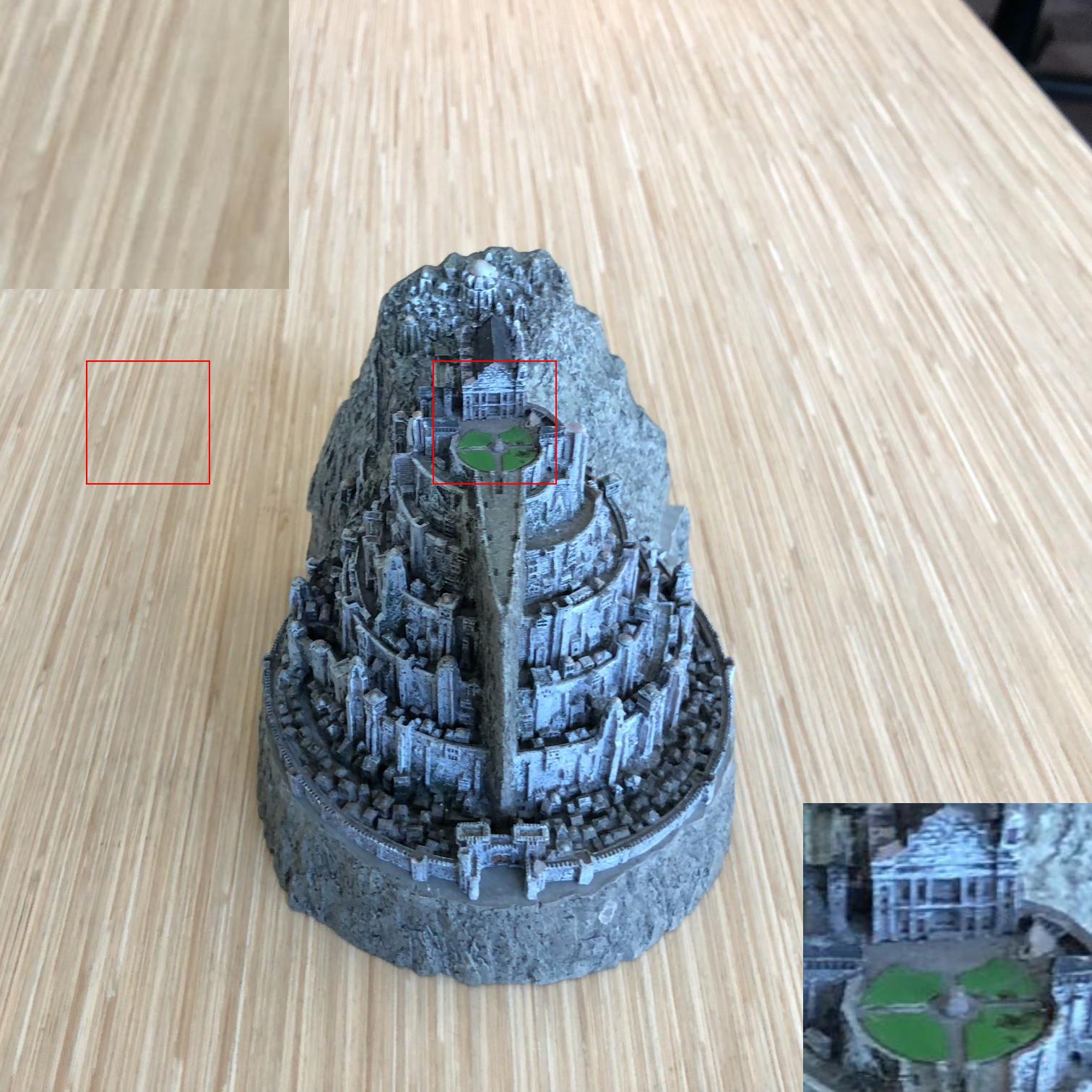} \\

    \includegraphics[width=0.21\textwidth,valign=m]{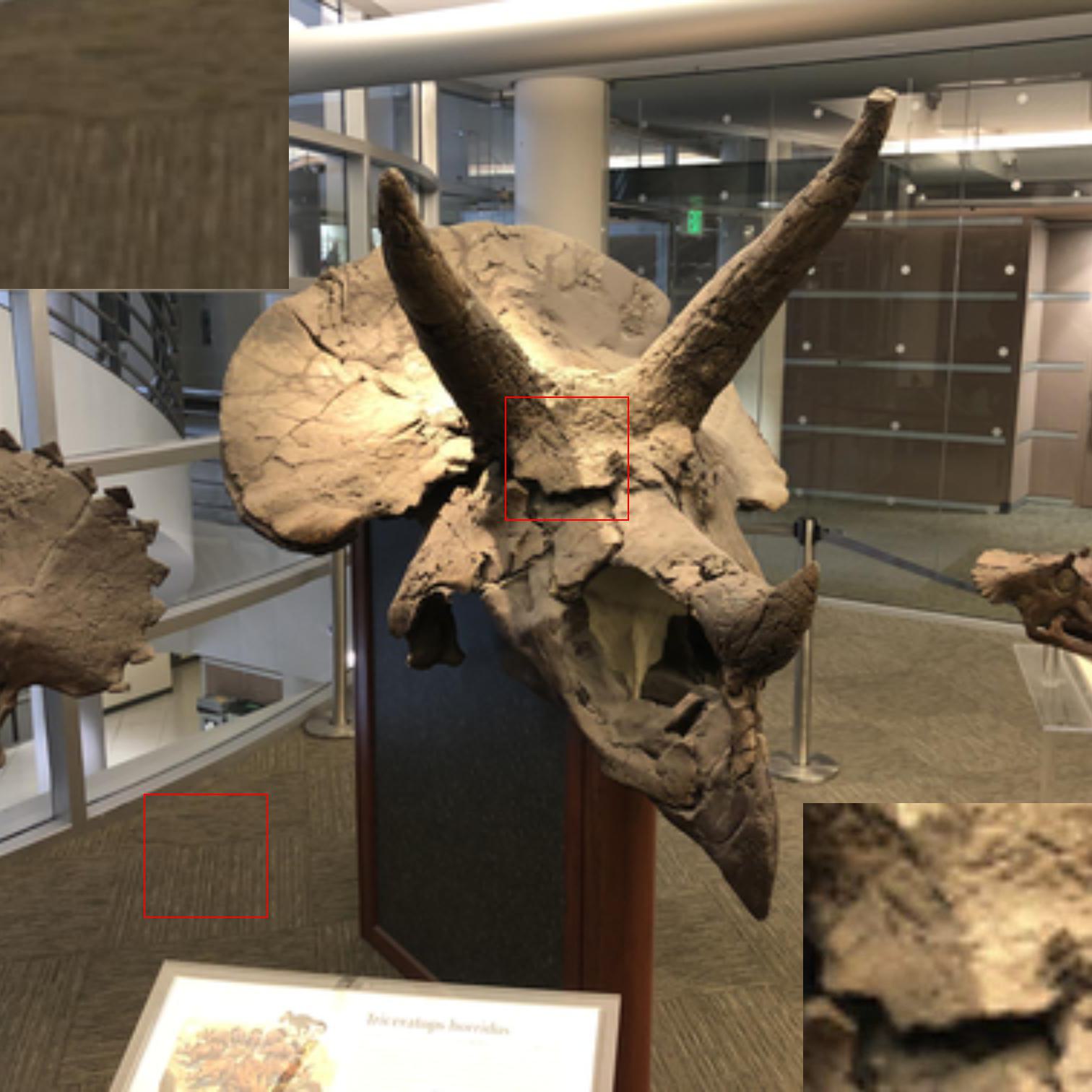} & 
    \includegraphics[width=0.21\textwidth,valign=m]{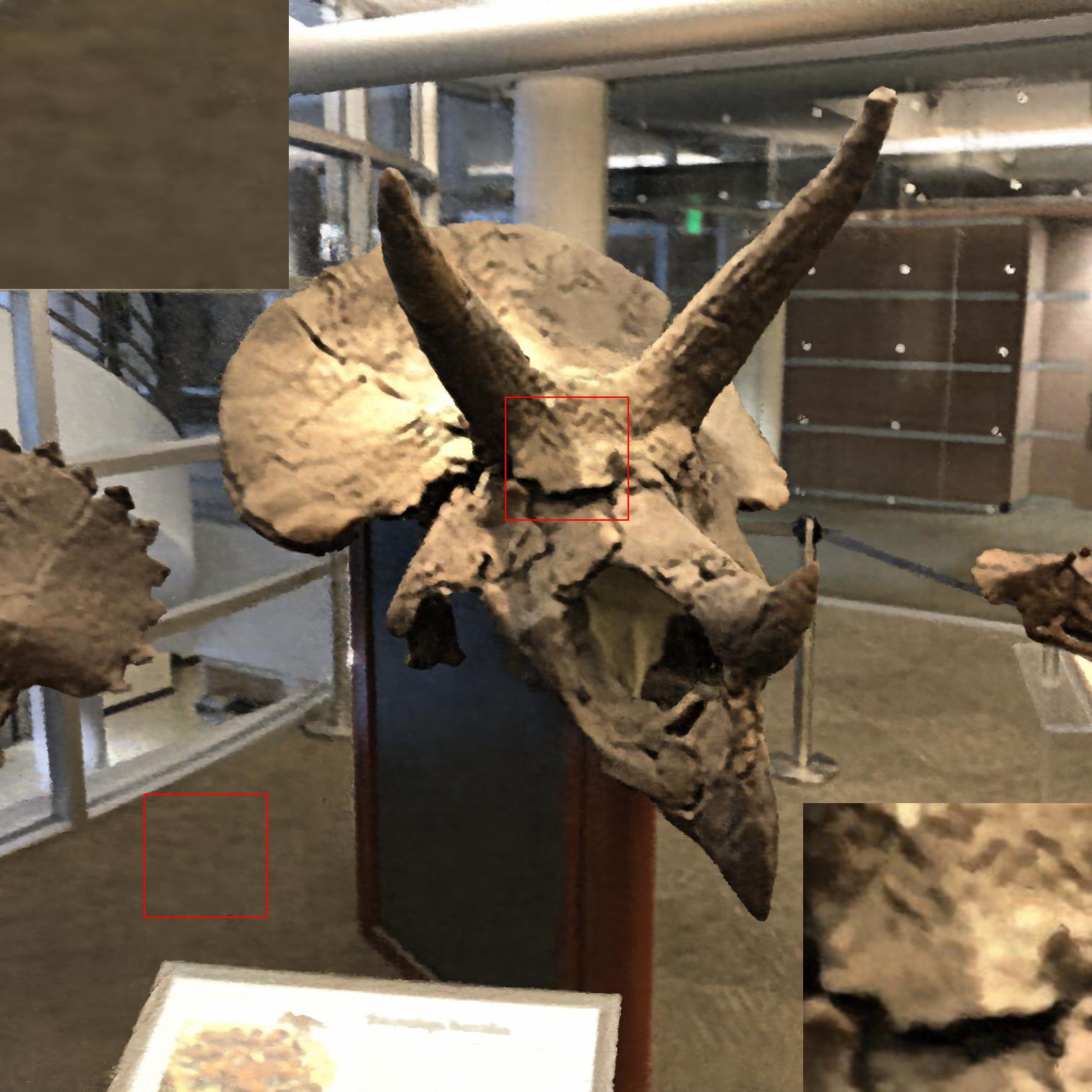} &
    \includegraphics[width=0.21\textwidth,valign=m]{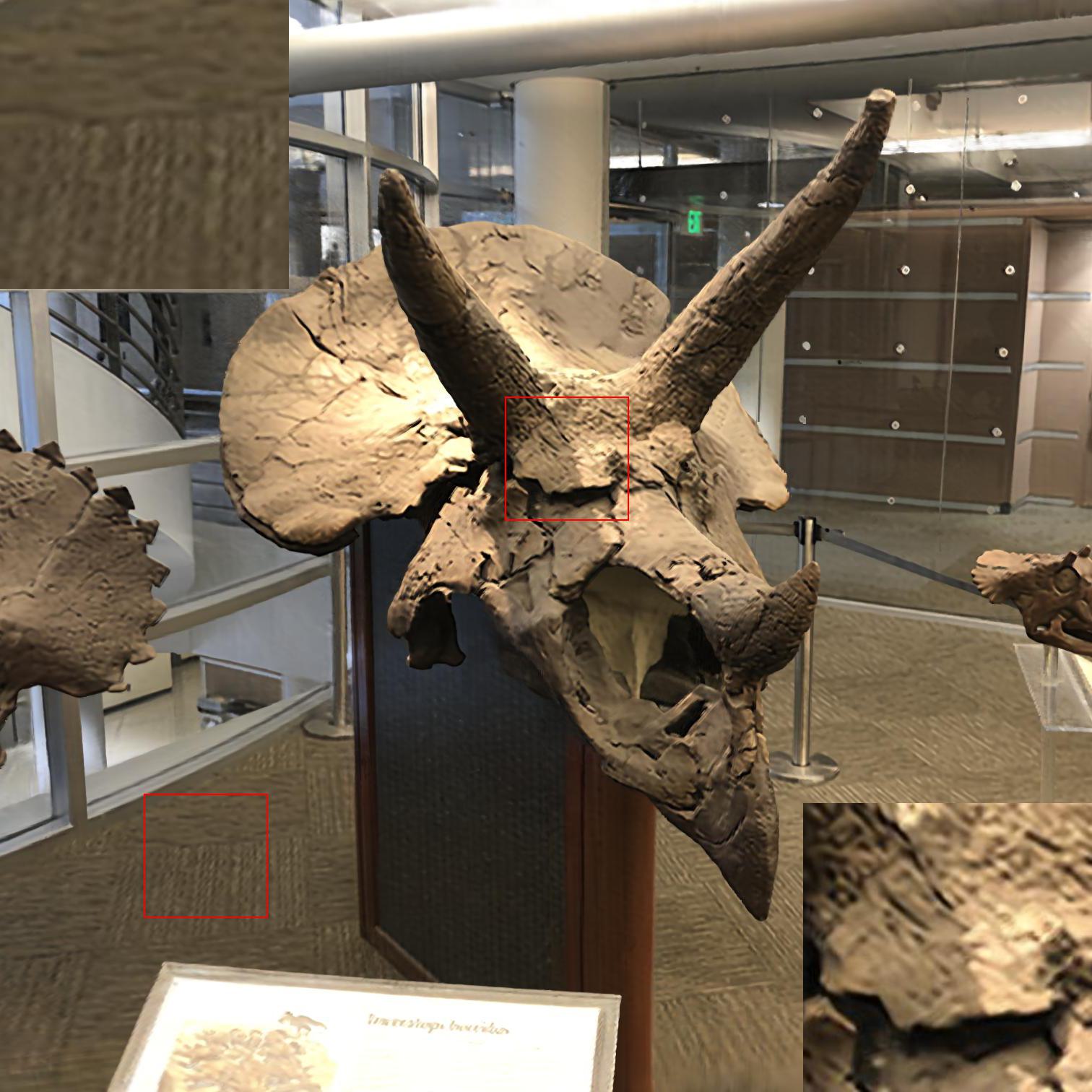} &
    \includegraphics[width=0.21\textwidth,valign=m]{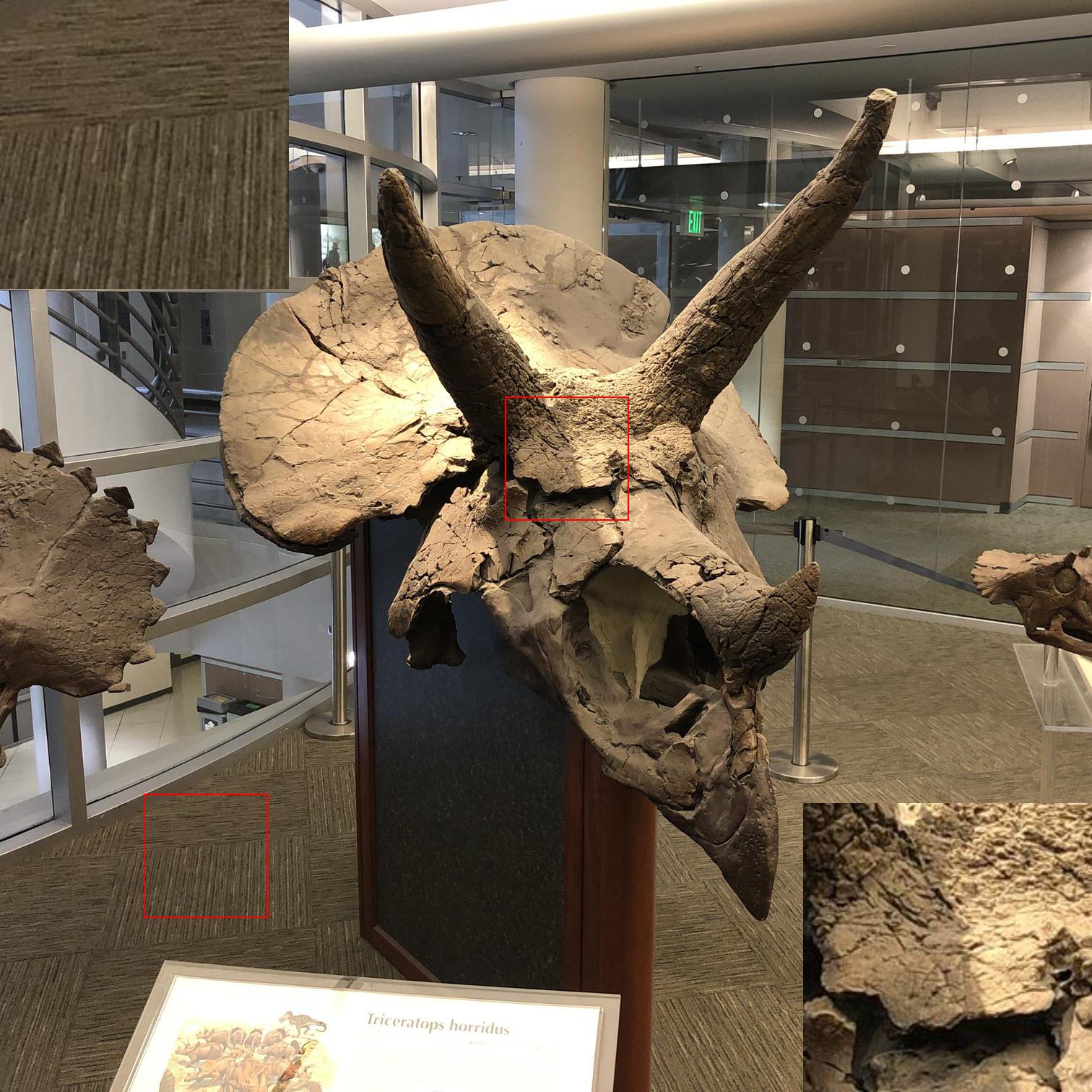} \\

    \includegraphics[width=0.21\textwidth,valign=m]{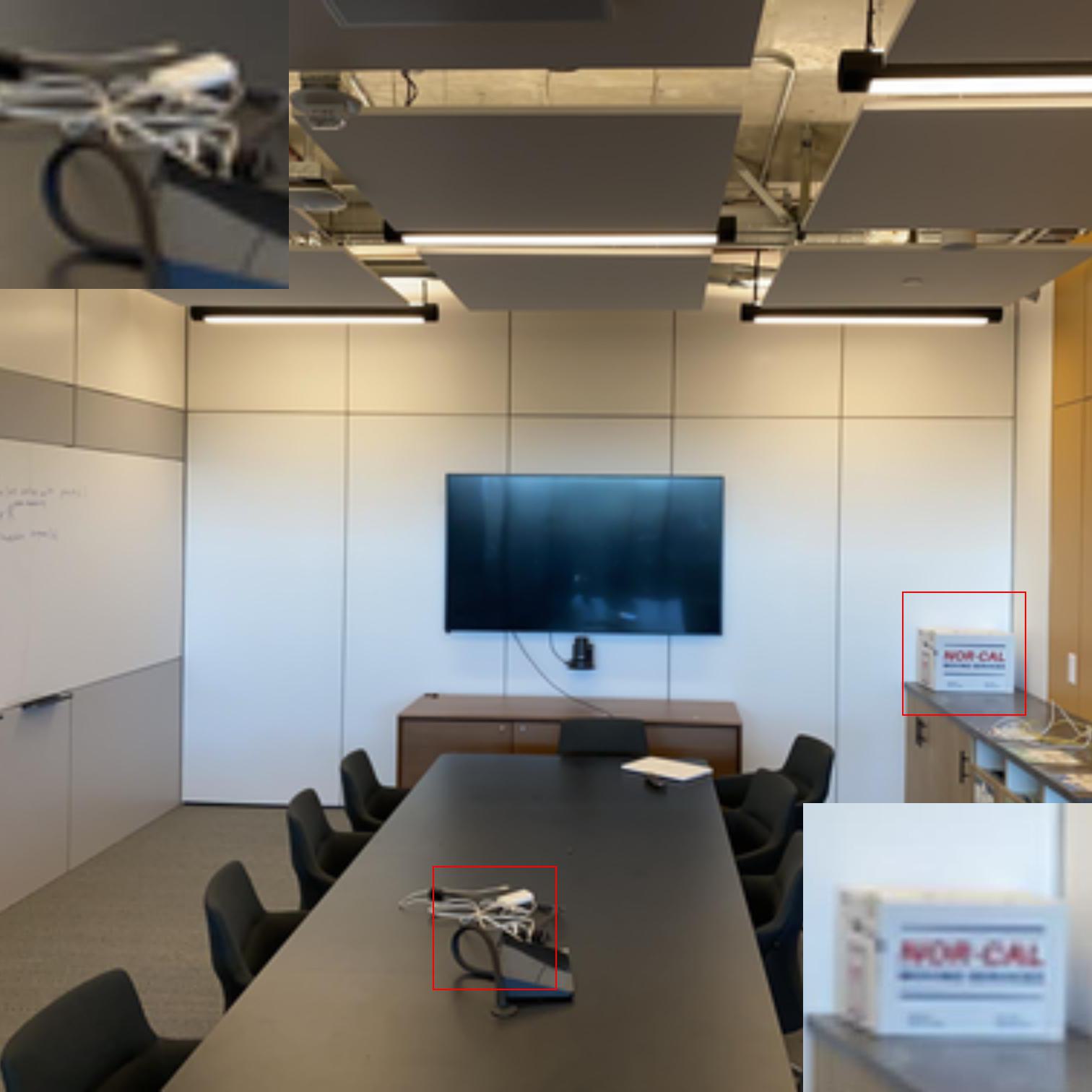} & 
    \includegraphics[width=0.21\textwidth,valign=m]{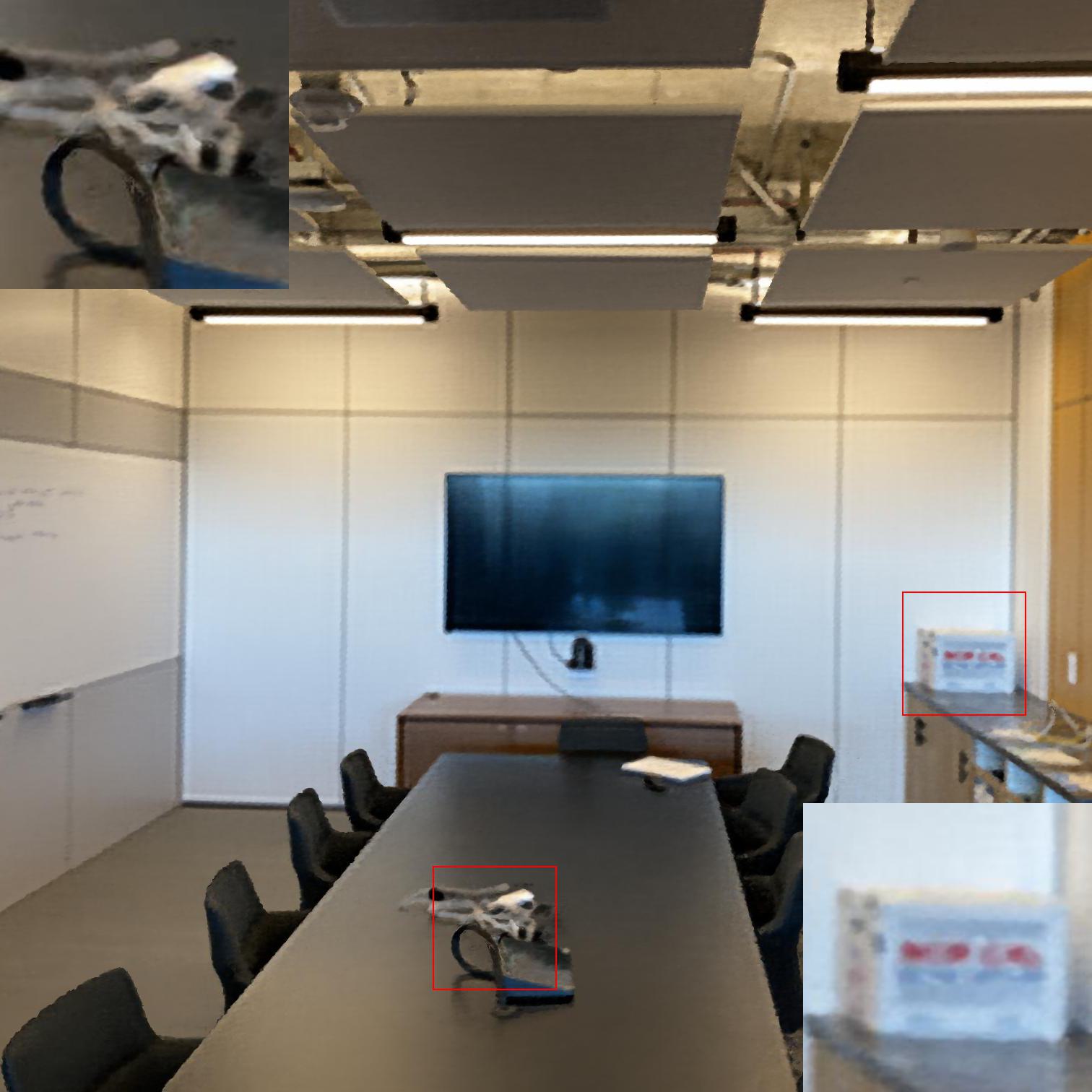} &
    \includegraphics[width=0.21\textwidth,valign=m]{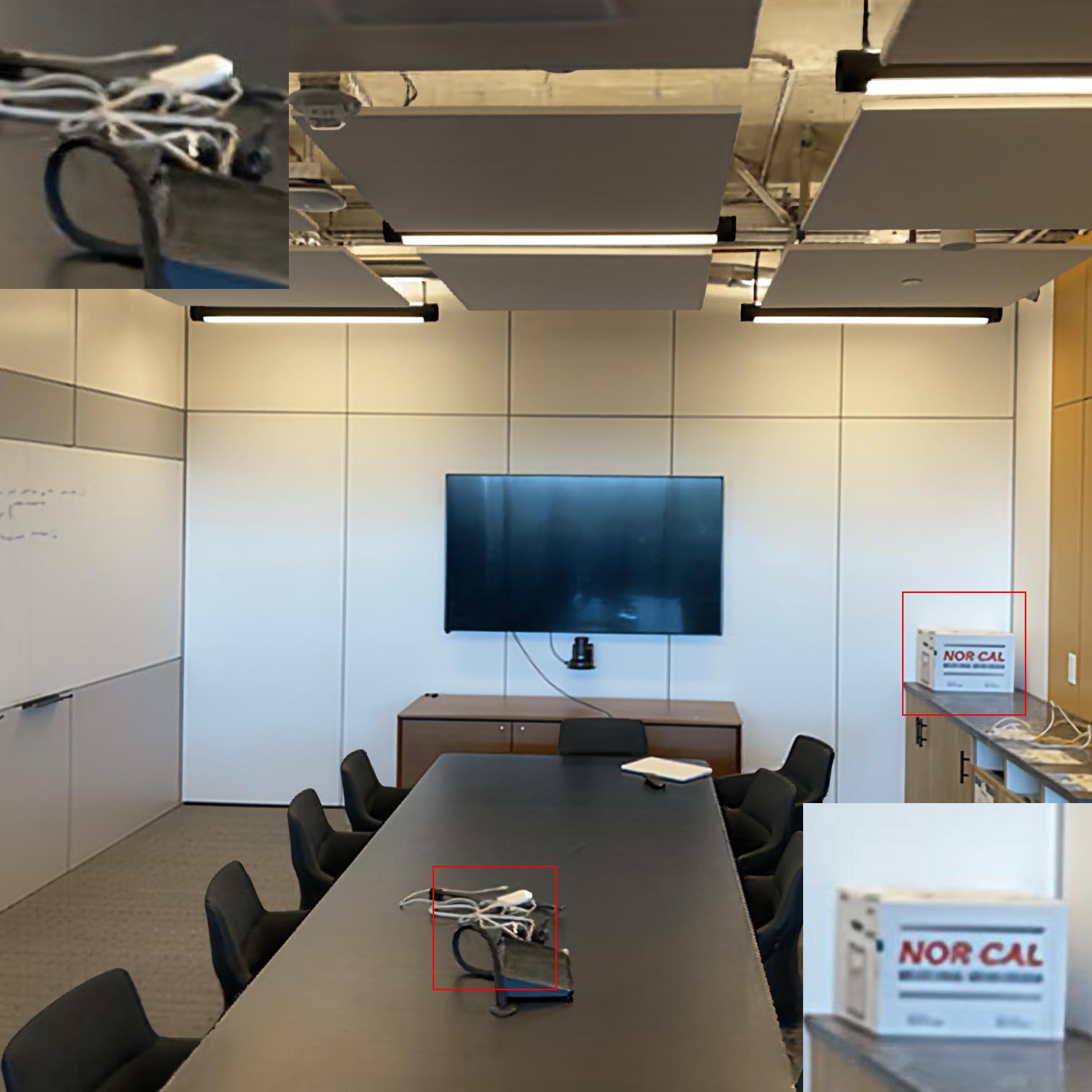} &
    \includegraphics[width=0.21\textwidth,valign=m]{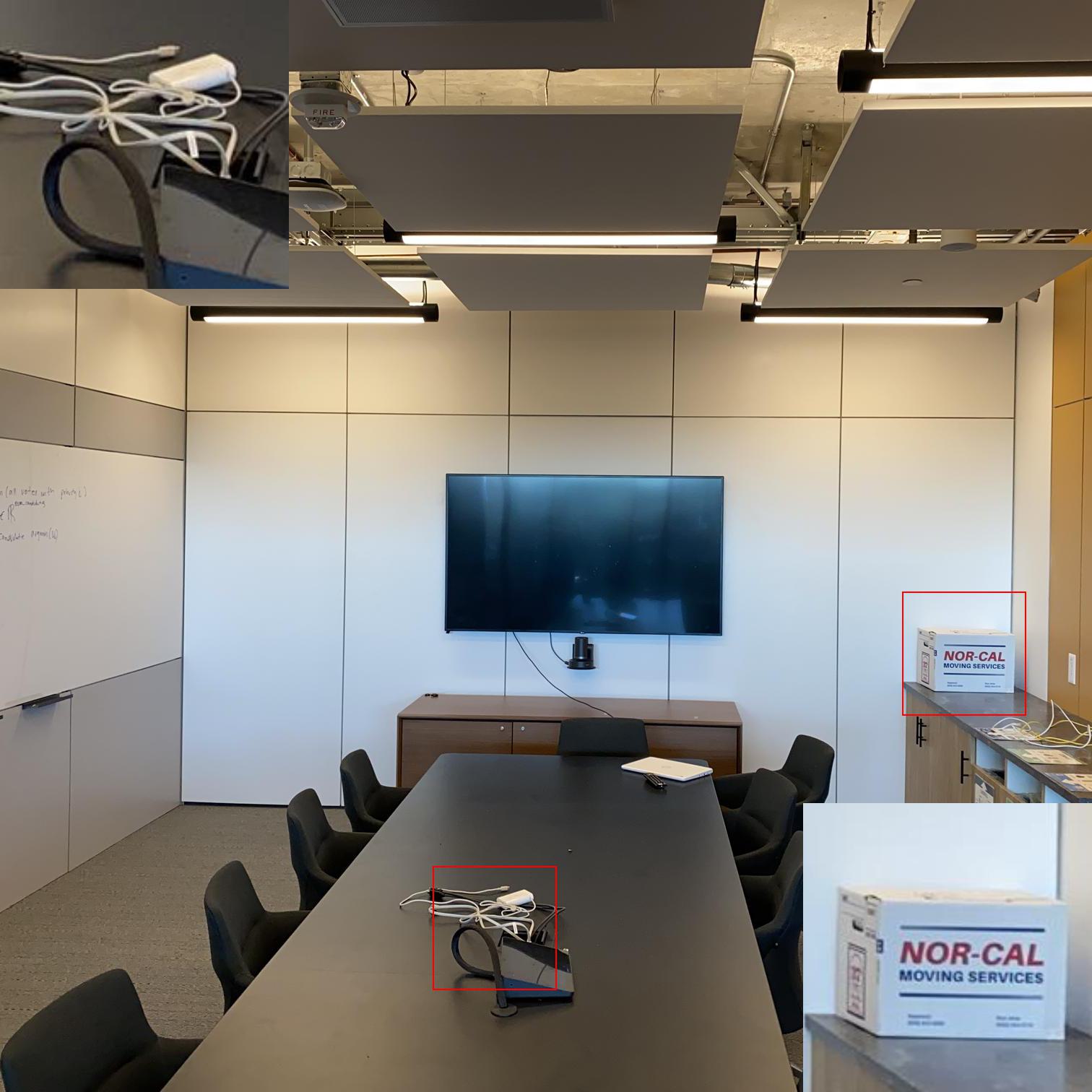} \\

    \end{tabular}}
    \caption{\emph{NeRF LLFF super-resolution qualitative results}.}
    \label{fig:qualitive_results_llff_sr}
\end{figure}

\begin{figure}[h!]
    \centering
    \begin{subfigure}[b]{0.8\columnwidth} 
        \centering
        \includegraphics[width=\textwidth]{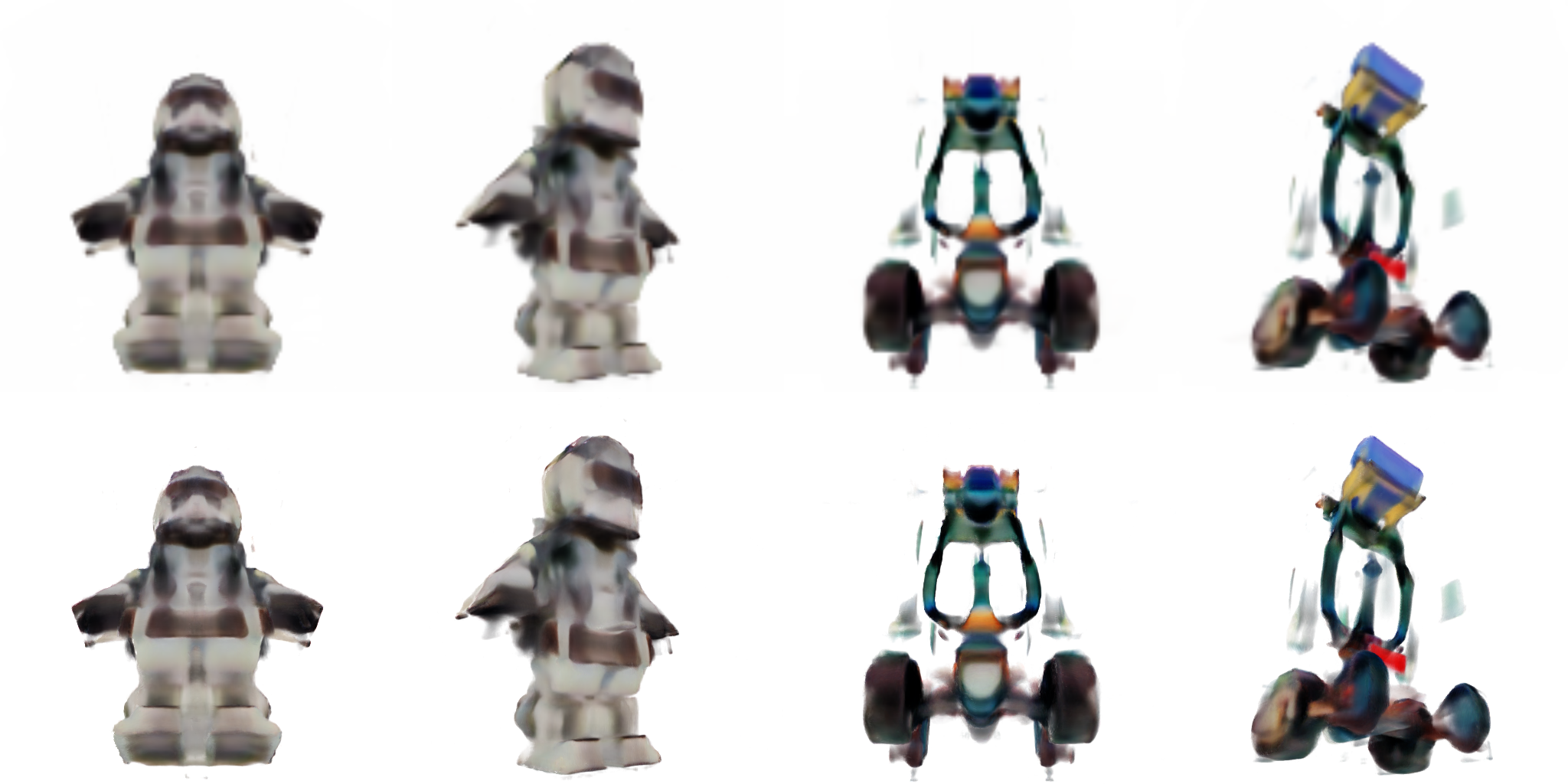}
    \end{subfigure}%
     \vspace{-0.1in}
     \caption{SR of low-res objects generated by triplane diffusion with text ``Astronaut suit and helmet" and ``Colorful electric scooter''. First row is low-res and second row is TriNerFLet-SR.}
      \vspace{-0.3in}
    \label{fig:sr_example_supp}
\end{figure}

Figures \ref{fig:qualitive_results_rec} and \ref{fig:qualitive_results_llff_sr} provide qualitative results for reconstruction and LLFF SR experiments respectively. For all results, the reader is referred to the project webpage \url{https://rajaeekh.github.io/trinerflet-web}.
Furthermore, TriNeRFLet-SR can be incorporated into a generative model. To show that, we use the diffusion process in \cite{hong20243dtopia} that generates a triplane and then apply SR to it using our scheme. Fig. \ref{fig:sr_example_supp} presents examples of
triplanes rendered with resolution $128$ that are upscaled to $512$ by TriNeRFLet SR.

\end{document}